\documentclass[journal,onecolumn]{IEEEtran}
\pdfoutput=1
\usepackage[latin1]{inputenc}
\usepackage{graphicx}

\usepackage[caption=false,font=footnotesize]{subfig}
\usepackage{verbatim}

\author{Matthew Anderson, Geng-Shen Fu, Ronald Phlypo, and T\"{u}lay~Adal{\i}}
\title{Independent Vector Analysis: Identification Conditions and Performance Bounds}

\usepackage{cite}

\usepackage[cmex10]{amsmath}
\usepackage{amsfonts}
\usepackage{amssymb}
\usepackage{amsthm}

\newtheorem{theorem}{Theorem}
\newtheorem{lemma}{Lemma}
\newtheorem{definition}{Definition}
\newcommand{\abs}[1]{\left| #1\right|}

\def\ci{\perp\!\!\!\perp} 
\newcommand{\mx}[1]{\ensuremath{\mathbf{#1}}} 
\newcommand{\mxgk}[1]{\ensuremath{\boldsymbol{#1}}} 
\newcommand{\mxelement}[3]{\ensuremath{\left[#1\right]_{#2,#3}}} 
\newcommand{\vect}[1]{\ensuremath{\mathbf{#1}}} 
\newcommand{\vectgk}[1]{\ensuremath{\boldsymbol{#1}}} 
\newcommand{\vectelement}[2]{\ensuremath{\left[#1\right]_{#2}}} 

\newcommand{\nat}{\ensuremath{\mathbb{N}}} 
\newcommand{\real}{\ensuremath{\mathbb{R}}} 
\newcommand{\Tpose}{\ensuremath{\mathsf{T}}} 
\newcommand{\trace}[1]{\ensuremath{\mathrm{tr}\left(#1\right)}} 

\newcommand{\diag}[1]{\ensuremath{\mathrm{diag}\left(#1\right)}} 
\newcommand{\Diag}[1]{\ensuremath{\mathrm{Diag}\left(#1\right)}} 
\newcommand{\vectorize}[1]{\ensuremath{\mathrm{vec}\left(#1\right)}} 
\newcommand{\vectorizesub}[2]{\ensuremath{\mathrm{vec}_{#2}\left(#1\right)}} 
\newcommand{\vectorizesubT}[2]{\ensuremath{\mathrm{vec}^\Tpose_{#2}\left(#1\right)}} 

\newcommand{\kron}{\ensuremath{\otimes}} 
\newcommand{\hadamard}{\ensuremath{\circ}} 
\newcommand{\hadamardd}{\ensuremath{\oslash}} 

\newcommand{\variance}[1]{\ensuremath{\mathrm{var}\left\{#1\right\}}}
\newcommand{\covariance}[1]{\ensuremath{\mathrm{cov}\left\{#1\right\}}}
\newcommand{\expect}[1]{\ensuremath{E\left\{#1\right\}}} 
\newcommand{\entropy}[1]{\ensuremath{\mathcal{H}\left\{#1\right\}}} 
\newcommand{\mutinfo}[1]{\ensuremath{\mathcal{I}\left\{#1\right\}}} 
\newcommand{\entropyr}[1]{\ensuremath{\mathcal{H}_r\left\{#1\right\}}} 
\newcommand{\mutinfor}[1]{\ensuremath{\mathcal{I}_r\left\{#1\right\}}} 
\newcommand{\naturale}[1]{\ensuremath{\exp\left({#1}\right)}} 

\newcommand{\pdfnorm}[2]{\ensuremath{\mathcal{N}\left(#1,#2\right)}} 

\usepackage{etoolbox}
\usepackage[acronym,nonumberlist,style=list, 
				section,
            sanitize={description=false},
           ]{glossaries}
           
\makeatletter           
\appto\newacronymhook{%
  \newbool{glo@\the\glslabeltok @usedonlyonce}
}
\patchcmd{\@gls@}{%
  \glsunset{#2}%
}{
  \ifglsused{#2}{%
    \write\@auxout{\global\setbool{glo@#2@usedonlyonce}{false}}%
  }{%
    \write\@auxout{\global\setbool{glo@#2@usedonlyonce}{true}}%
  }%
  \glsunset{#2}%
}{}{}

\patchcmd{\@gls@}{%
  \glsentryfirst{#2}%
}{
  \ifbool{glo@#2@usedonlyonce}{\glsentrylong{#2}}{\glsentryfirst{#2}}%
}{}{}

\patchcmd{\@Gls@}{%
  \glsunset{#2}%
}{
  \ifglsused{#2}{%
    \write\@auxout{\global\setbool{glo@#2@usedonlyonce}{false}}%
  }{%
    \write\@auxout{\global\setbool{glo@#2@usedonlyonce}{true}}%
  }%
  \glsunset{#2}%
}{}{}

\patchcmd{\@Gls@}{%
  \glsentryfirst{#2}%
}{
  \ifbool{glo@#2@usedonlyonce}{\glsentrylong{#2}}{\glsentryfirst{#2}}%
}{}{}

\let\old@do@wrglossary\@do@wrglossary
\renewcommand{\@do@wrglossary}[1]{\ifbool{glo@#1@usedonlyonce}{}{\old@do@wrglossary{#1}}}

\makeatother

\newcommand{\hiac}[1]{{#1}} 

\newacronym[description={\hiac{n}onnegative \hiac{m}atrix \hiac{f}actorization}] 
{nmf}{\mbox{NMF}}{nonnegative matrix factorization} 
\newacronym[description={\hiac{b}lind \hiac{s}ource \hiac{s}eparation}] 
{bss}{\mbox{BSS}}{blind source separation} 
\newacronym[description={\hiac{j}oint \hiac{b}lind \hiac{s}ource \hiac{s}eparation}]
{jbss}{\mbox{JBSS}}{joint blind source separation}
\newacronym[description={\hiac{c}anonical \hiac{c}orrelation \hiac{a}nalysis, see \cite{Hotelling_CCA_1936}}]
{cca}{\mbox{CCA}}{canonical correlation analysis}
\newacronym[description={\hiac{m}ultiset \hiac{c}anonical \hiac{c}orrelation \hiac{a}nalysis, see
\cite{Kettenring_MCCA_1971,Li_MCCA_2009}}]
{mcca}{\mbox{MCCA}}{multiset canonical correlation analysis}
\newacronym[description={\hiac{i}ndependent \hiac{c}omponent \hiac{a}nalysis}]
{ica}{\mbox{ICA}}{independent component analysis}
\newacronym[description={\hiac{fastICA} algorithm, see \cite{Hyvarinen_fastica_1999}}]
{fica}{\mbox{FastICA}}{fast independent component analysis}
\newacronym[description={decoupled independent component analysis algorithm, see \cite{Anderson_DICA_2012}}]
{dica}{\mbox{D-ICA}}{decoupled ICA}
\newacronym[description={\hiac{m}ultidimensional \hiac{i}ndependent \hiac{c}omponent  \hiac{a}nalysis, see \cite{Cardoso_MICA_1998}}]
{mica}{\mbox{MICA}}{multidimensional independent component  analysis}
\newacronym[description={\hiac{e}fficient \hiac{f}ast \hiac{i}ndependent \hiac{c}omponent \hiac{a}nalysis algorithm, see \cite{Koldovsky_EFICA_2006}}]
{efica}{\mbox{EFICA}}{efficient fast independent component analysis}
\newacronym[description={\hiac{o}ptimally \hiac{a}ligned \hiac{EFICA} using clairvoyant knowldege}]
{oa-efica}{\mbox{OA-EFICA}}{optimally aligned \hiac{EFICA}}
\newacronym[description={\hiac{a}lgorithm for \hiac{m}ultiple \hiac{u}nknown \hiac{s}ignals \hiac{e}xtraction of \cite{Tong_AMUSE_1990}}]
{amuse}{\mbox{AMUSE}}{algorithm for multiple unknown signals extraction}
\newacronym[description={\hiac{s}econd-\hiac{o}rder \hiac{b}lind \hiac{i}dentification by joint diagonalization of temporal correlation matrices, see \cite{Belouchrani_SOBI_1993}}]
{sobi}{\mbox{SOBI}}{second-order blind identification}
\newacronym[description={\hiac{w}eights-\hiac{a}djusted \hiac{s}econd-\hiac{o}rder \hiac{b}lind \hiac{i}dentification by joint diagonalization of temporal correlation matrices, see \cite{Yeredor_Wasobi_2000}}]
{wasobi}{\mbox{WASOBI}}{weights-adjusted second-order blind identification}
\newacronym[description={\hiac{s}trongly \hiac{u}ncorrelated \hiac{t}ransform, see \cite{Eriksson_SUT_2004}}]
{sut}{\mbox{SUT}}{strongly uncorrelated transform}
\newacronym[description={\hiac{e}ntropy \hiac{b}ound \hiac{m}inimization, see \cite{Li_EBM_2009,Li_RealEBM_2010}}]
{ebm}{\mbox{EBM}}{entropy bound minimization}
\newacronym[description={\hiac{e}ntropy \hiac{r}ate \hiac{b}ound \hiac{m}inimization, see \cite{XiLinLi_FBSS_2010}}]
{erbm}{\mbox{ERBM}}{entropy rate bound minimization}

\newacronym[description={\hiac{i}ndependent \hiac{v}ector \hiac{a}nalysis}]
{iva}{\mbox{IVA}}{independent vector analysis}
\newacronym[description={\hiac{j}oint \hiac{diag}onalization via \hiac{s}econd-\hiac{o}rder \hiac{s}tatistics, see \cite{XiLinLi_JointDiagGradient_2010,XiLinLi_JointDiagOrthoProcustes_2011}}]
{jdiag-sos}{\mbox{JDIAG-SOS}}{joint diagonalization via second-order statistics}
\newacronym[description={\hiac{j}oint \hiac{diag}onalization via \hiac{cum}ulants of fourth-order}]
{jdiag-cum4}{\mbox{JDIAG-CUM4}}{joint diagonalization via cumulants of fourth-order}
\newacronym[description={\hiac{IVA} with second-order uncorrelated multivariate \hiac{L}aplace distribution model}]
{iva-l}{\mbox{IVA-Lap}}{IVA with second-order uncorrelated multivariate Laplace distribution model}
\newacronym[description={\hiac{IVA} with multivariate \hiac{G}aussian distribution model}]
{iva-g}{\mbox{IVA-Gauss}}{IVA with multivariate Gaussian distribution model}
\newacronym[description={\hiac{IVA-G} cost function optimize using \hiac{v}ector gradient descent approach}]
{iva-g-v}{\mbox{IVA-G-V}}{IVA-G cost function optimize using vector gradient descent approach}
\newacronym[description={\hiac{IVA-G} cost function optimize using \hiac{N}ewton algorithm}]
{iva-g-n}{\mbox{IVA-G-N}}{IVA-G cost function optimize using Newton algorithm}
\newacronym[description={\hiac{IVA-G} cost function optimize using \hiac{b}lock quasi-Newton algorithm}]
{iva-g-b}{\mbox{IVA-G-B}}{IVA-G cost function optimize using block quasi-Newton algorithm}
\newacronym[description={\hiac{IVA-G} cost function optimize using quasi-Newton algorithm}]
{iva-g-q}{\mbox{IVA-G-Q}}{IVA-G cost function optimize using quasi-Newton algorithm}
\newacronym[description={\hiac{IVA} algorithm using \protect\glsentrytext{iva-g} solution to initialize  \protect\glsentrytext{iva-l}}]
{iva-gl}{\mbox{IVA-GL}}{IVA algorithm using \protect\gls{iva-g} solution to initialize \protect\gls{iva-l}}
\newacronym[description={\hiac{IVA} with multivariate \hiac{K}otz distribution model}]
{iva-kotz}{\mbox{IVA-Kotz}}{IVA with multivariate Kotz distribution model}
\newacronym[description={\hiac{IVA} with \hiac{m}ultivariate \hiac{p}ower \hiac{e}xponential distribution model}]
{iva-mpe}{\mbox{IVA-MPE}}{IVA with multivariate power exponential distribution model}
\newacronym[description={\hiac{IVA} with \hiac{n}on-\hiac{c}ircular multivariate \hiac{G}aussian distribution model}]
{iva-nc-g}{\mbox{IVA-NC-G}}{IVA with noncircular multivariate Gaussian distribution model}
\newacronym[description={\hiac{IVA-NC-G} cost function optimize using \hiac{v}ector gradient descent approach}]
{iva-nc-g-v}{\mbox{IVA-NC-G-V}}{IVA-NC-G cost function optimize using vector gradient descent approach}
\newacronym[description={\hiac{IVA-NC-G} cost function optimize using \hiac{N}ewton algorithm}]
{iva-nc-g-n}{\mbox{IVA-NC-G-N}}{IVA-NC-G cost function optimize using Newton algorithm}
\newacronym[description={\hiac{IVA-NC-G} cost function optimize using quasi-Newton algorithm}]
{iva-nc-g-q}{\mbox{IVA-NC-G-Q}}{IVA-NC-G cost function optimize using quasi-Newton algorithm}

\newacronym[description={\hiac{m}agnetic \hiac{r}esonance \hiac{i}maging}]
{mri}{\mbox{MRI}}{magnetic resonance imaging}
\newacronym[description={\hiac{f}unctional \hiac{m}agnetic \hiac{r}esonance \hiac{i}maging}]
{fmri}{\mbox{fMRI}}{functional magnetic resonance imaging}
\newacronym[description={\hiac{b}lood \hiac{o}xygenation \hiac{l}evel \hiac{d}ependent}]
{bold}{\mbox{BOLD}}{blood oxygenation level dependent}
\newacronym[description={\hiac{g}eneral \hiac{l}inear \hiac{m}odel}]
{glm}{\mbox{GLM}}{general linear model}
\newacronym[description={\hiac{s}tructural \hiac{m}agnetic \hiac{r}esonance \hiac{i}maging}]
{smri}{\mbox{sMRI}}{structural magnetic resonance imaging}
\newacronym[description={\hiac{e}lectro\hiac{e}ncephalo\hiac{g}raphy}]
{eeg}{\mbox{EEG}}{electroencephalography}
\newacronym[description={\hiac{e}lectro\hiac{c}ardio\hiac{g}ram}]
{ecg}{\mbox{ECG}}{electrocardiogram}

\newacronym[description={\hiac{p}robability \hiac{d}istribution \hiac{f}unction}]
{pdf}{pdf}{probability distribution function}
\newacronym[description={\hiac{c}umulative \hiac{d}istribution \hiac{f}unction}]
{cdf}{cdf}{cumulative distribution function}
\newacronym[description={\hiac{F}isher \hiac{i}nformation \hiac{m}atrix}]
{fim}{\mbox{FIM}}{Fisher information matrix}
\newacronym[description={\hiac{C}ram\'{e}r-\hiac{R}ao \hiac{l}ower \hiac{b}ound}]
{crlb}{\mbox{CRLB}}{Cram\'{e}r-Rao lower bound}
\newacronym[description={\hiac{i}nduced \hiac{C}ram\'{e}r-\hiac{R}ao \hiac{l}ower \hiac{b}ound}]
{icrlb}{\mbox{iCRLB}}{induced Cram\'{e}r-Rao lower bound}
\newacronym[description={\hiac{i}nterference to \hiac{s}ource \hiac{r}atio}]
{isr}{\mbox{ISR}}{interference to source ratio}
\newacronym[description={\hiac{i}nter\hiac{s}ymbol-\hiac{i}nterference}]
{isi}{\mbox{ISI}}{intersymbol-interference}
\newacronym[description={\hiac{s}ignal to \hiac{n}oise \hiac{r}atio}]
{snr}{\mbox{SNR}}{signal to noise ratio}
\newacronym[description={\hiac{m}aximum \hiac{l}ikelihood \hiac{e}stimate}]
{mle}{\mbox{MLE}}{maximum likelihood estimate}
\newacronym[description={\hiac{m}ultivariate \hiac{p}ower \hiac{e}xponential distribution, also known as \glsentrytext{mggd}}]
{mpe}{\mbox{MPE}}{multivariate power exponential}
\newacronym[description={\hiac{m}ultivariate \hiac{g}eneralized \hiac{G}aussian \hiac{d}istribution, also known as \glsentrytext{mpe}}]
{mggd}{\mbox{MGGD}}{multivariate generalized Gaussian distribution}

\newacronym[description={\hiac{v}ector \hiac{a}uto\hiac{r}egressive model of order $p$}]
{var}{$\text{VAR}\left(p\right)$}{vector autoregressive model of order $p$}
\newacronym[description={\hiac{a}uto\hiac{r}egressive model of order $p$}]
{ar}{$\text{AR}\left(p\right)$}{autoregressive model of order $p$}

\newacronym[description={\hiac{s}ource \hiac{c}omponent \hiac{v}ector, i.e., $\mathbf{s}_{n}=\left[s_{n}^{[1]}, \ldots, s_{n}^{[K]}\right]^\Tpose$}]
{scv}{\mbox{SCV}}{source component vector}
\newacronym[description={\hiac{s}ource \hiac{c}omponent \hiac{m}atrix, i.e., $\mx{S}_{n}=\left[\vect{s}_{n}^{[1]}, \ldots, \vect{s}_{n}^{[K]}\right]^\Tpose$}]
{scm}{\mbox{SCM}}{source component matrix}
\newacronym[description={\hiac{g}eneralized \hiac{G}aussian \hiac{d}istribution, also known as generalized normal distribution}]
{ggd}{\mbox{GGD}}{generalized Gaussian distribution}
\newacronym[description={\hiac{n}onorthogonal \hiac{d}ecoupling \hiac{t}rick}]
{ndt}{\mbox{NDT}}{nonorthogonal decoupling trick}

\newacronym[description={\hiac{w}ith \hiac{r}espect \hiac{t}o}]
{wrt}{wrt}{with respect to}

\newacronym[description={\hiac{i}ndependently and \hiac{i}dentically \hiac{d}istributed}]
{iid}{iid}{independently and identically distributed}

\newacronym[description={\hiac{if} and only \hiac{i}f}]
{iff}{iff}{if and only if}

\newcommand{\figref}[1]{{Fig.~\ref{#1}}} 

\begin{document}

\maketitle

\begin{abstract}
Recently, an extension of \gls{ica} from one to multiple datasets, termed \gls{iva}, has been the subject of significant research interest.  
\Gls{iva} has also been shown to be a generalization of Hotelling's canonical correlation analysis.
In this paper, we provide the identification conditions for a general \gls{iva} formulation, which accounts for linear, nonlinear, and  sample-to-sample dependencies.   
The identification conditions are a generalization of previous results for \gls{ica} and for \gls{iva} when samples are \glsdesc{iid}.  
Furthermore,  a principal aim of \gls{iva} is the identification of dependent sources between datasets.
Thus, we provide the additional conditions for when the arbitrary ordering of the sources within each dataset is common.
Performance bounds in terms of the \glsdesc{crlb} are also provided for the demixing matrices and \glsdesc{isr}.
The performance of two \gls{iva} algorithms are compared to the theoretical bounds.
\end{abstract}

\glsresetall

\section{Motivation and  Introduction}
\Gls{bss} problems have been well studied and many algorithms have been developed and successfully applied in a vast array of applications \cite{Ica_Book_2001,Comon_HandbookBSS_2010}.  
A generalization of the \gls{bss} problem to multiple datasets, termed \gls{jbss}, has been introduced recently \cite{Lee_IVAfMRI_2008,Li_MCCA_2009}.  
The recent interest in \gls{jbss} is motivated by various application domains such as when analyzing multisubject datasets in biomedical studies using \gls{fmri} or \gls{eeg} data \cite{Lee_IVAfMRI_2008, Li_MCCA_2009} or when solving the convolutive \gls{ica} problem in the frequency domain using multiple frequency bins \cite{Kim_RealtimeIVA_2010}.
Interestingly, several algorithms developed prior to the development of the \gls{bss} concept are capable of achieving \gls{jbss} \cite{Kettenring_MCCA_1971,Young_NonlinearCCA_1976}.
Thus, a much larger set of applications than the examples above are well treated using the \gls{jbss} formulation.

One particular formulation of \gls{jbss} has been termed \gls{iva}. 
The formulation of \gls{iva} is an extension of the (linear, instantaneous) \gls{ica} model.  \Gls{iva} assumes a source within one dataset is dependent on at most one source in another dataset while 
sources within a dataset are mutually independent (as in \gls{ica}).
Thus, \gls{iva} reduces to performing \gls{ica} individually on each dataset when sources possess no dependence across datasets.
Of particular interest here is to determine the conditions when \gls{iva} is identifiable.  
For a real-valued single dataset problem, independent sources can be `blindly' identified up to a permutation and scaling ambiguity as long as no two sources are Gaussian with proportional sample-to-sample correlation matrices \cite[Chapter 4]{Comon_HandbookBSS_2010}. 
The \gls{iva} framework has been shown to possess an additional type of diversity which can be exploited for identifying sources that cannot be identified by \gls{ica}, \cite{Anderson_IVAG_2012}. 

In this paper, a general framework for \gls{iva} is presented. 
By `general' we mean an \gls{iva} formulation that accounts for dependency between samples, i.e., when the samples are not  \gls{iid}.
Prior to introducing this \gls{iva} formulation in Section IV, we give a review of existing \gls{iva} algorithms in Section II and define our mathematical conventions and notations in Section III.
Naturally, \gls{iva} can be achieved by maximizing the likelihood function, which is shown in Section V to be the same in practice as minimizing the entropy rate (subject to a regularity term).
The likelihood function has an associated \gls{fim} of a form that we describe in Section VI.  
The \gls{fim} is used in deriving the identification conditions and source separation performance bounds in Sections VII and VIII, respectively.
The \gls{iva} identification conditions and performance bounds are generalizations of the results for \gls{ica} (of a single dataset).
The \gls{iva} case when samples are \gls{iid} is shown to have a performance bound that can be expressed compactly for the very large class of multivariate elliptical distributions.
In Section IX, the performance bounds are compared to the performance achieved by two previously published algorithms for \gls{iva}.
In the last section, we discuss directions for future work.

\section{Review of Existing \glsentrytext{iva} Algorithms}
As mentioned previously, the origins of algorithms that can be used for \gls{iva} date back to pre-\gls{ica} times.  
In fact, classical \gls{cca} \cite{Hotelling_CCA_1936}  achieves \gls{iva} for linearly dependent sources in analysis of two datasets.  
The formulation of \gls{cca} can be shown to serve as a basis for all \gls{iva} algorithms reviewed here.  
This is because \gls{cca} can be derived from two different, but related principles; maximum likelihood and eigenanalysis (diagonalization).
Here, we choose to separate the approaches into three classes for our review based on the source diversity exploited to achieve \gls{jbss}.  
It will be shown that each type of diversity can be utilized---independent of the other two---to achieve \gls{iva}.

\subsection{Linear dependence}
The first class is applicable to problems in which the sources are assumed to have linear dependence across datasets, but are linearly independent within datasets. 
The earliest approaches to extending \gls{cca} beyond two datasets are summarized in \cite{Kettenring_MCCA_1971} and has been termed \gls{mcca} in \cite{Li_MCCA_2009}.  The approaches within \gls{mcca} use cost functions based on second-order statistics that result in \gls{jbss} solutions that can be widely applied. 
Another approach to \gls{jbss} for linearly dependent sources can be derived using equivalently maximum likelihood or minimization of mutual information and results in \gls{iva-g} \cite{Anderson_IVAMGconf_2010, Via_ML_IVA_MLSP2011}. 

Since \gls{cca} can be achieved using generalized eigenvalue decomposition, it can also be posed as a diagonalization problem, which can be readily extended to achieve \gls{iva} using `generalized joint diagonalization'  \cite{XiLinLi_JointDiagOrthoProcustes_2011}.  For \gls{iva} of linearly dependent sources the covariance and cross-covariance matrices among the estimated sources in each dataset can be diagonalized as in \cite{XiLinLi_JointDiagGradient_2010,XiLinLi_JointDiagOrthoProcustes_2011}.
\subsection{Nonlinear dependence}
When the sources possess nonlinear dependence across the datasets then higher-order statistics should be utilized either explicitly or implicitly. 
The extension of \gls{cca} to nonlinear dependence measures for two datasets dates back to at least 1976 \cite{Young_NonlinearCCA_1976}.  Extensions to multiple datasets is given in \cite{deLeeuw_GifiSystem_1984}.
These early works are summarized in \cite{Gifi_Book_1990}. 

Another extension for nonlinear \gls{cca} of two datasets uses nonparametric univariate and bivariate density estimators in order to maximize the mutual information between two canonical correlation variates  \cite{Yin_ICCA_2004}.  
Kernels have also been used to transform the random vectors into a `feature-space' where linear \gls{cca} is then applied \cite{Akaho_KernelCCA_2001,Melzer_NonlinearCCA_2001}.  
A different type of transformation is proposed in \cite{Todros_MTCCA_2012}.  
Here measure transform functions are specified for transforming joint probability measures to identify nonlinearly dependent sources. 
To use either the kernel or measure transform approaches, one must determine the appropriate transform and transform parameters to achieve \gls{jbss} for the problem at hand.

\Gls{iva} also provides a framework for exploiting nonlinear dependencies.  
\Gls{iva}, as first introduced in \cite{Kim_IVA_2006,Kim_IVAasilomar_2006} and in the similar work of \cite{Hiroe_AlmostIVA_2006}, extends \gls{ica} to multiple datasets so as to solve the permutation ambiguity problem associated with frequency domain \gls{ica} \cite{Smaragdis_ConvolvedBSS_1998}.  
The nonlinear dependencies can be accounted for within the \gls{iva} framework by considering non-Gaussian sources.
For example, in \cite{Kim_IVA_2006,Kim_IVAasilomar_2006}, a nonlinear score function consistent with the second-order uncorrelated multivariate Laplacian distribution is used.

As is the case for linear dependence, diagonalization methods for \gls{iva} of nonlinearly dependent sources can be utilized.
Specifically, demixing matrices that diagonalize the higher-order statistics (i.e., cumulants of order higher than two) associated with the estimated sources are found \cite{XiLinLi_JointDiagGradient_2010,XiLinLi_JointDiagOrthoProcustes_2011,Phlypo_CDA_2013}.

\subsection{Sample-to-sample dependence}
Naturally for \gls{iva}, as for \gls{ica}, algorithms can be developed to exploit sample-to-sample dependence.  A \emph{generalization} of joint diagonalization provides such a solution by 
sampling the vector autocorrelation function at different time lags and finding demixing matrices which minimize correlation between the sources for all time lags, see, e.g., \cite{XiLinLi_JointDiagGradient_2010, XiLinLi_JointDiagOrthoProcustes_2011}.

\section{Mathematical Preliminaries}
For this paper, the domains are restricted to the sets of real (\real) and nonnegative natural (\nat) numbers.  Matrices and vectors from each domain are indicated by $\real^{M \times N}$, $\real^{M}$, $\nat^{M \times N}$, and $\nat^{M}$, respectively.
Scalar, (column) vector, and matrix quantities are denoted as lower-case light face, lower-case bold face, and upper-case bold face, respectively. 
The $m$th element of a vector $\vect{v}$, $\vectelement{\vect{v}}{m}$, and an element in the $m$th row and $n$th column of a matrix $\mx{A}$, $\mxelement{\mx{A}}{m}{n}$, are often denoted $v_{m}$ and $a_{m,n}$, respectively.

The Kronecker delta, $\delta_{m,n}$, is one when $m=n$ and zero otherwise. 
The standard basis vector, $\vect{e}_n$, is the the $n$th column of identity matrix, $\mx{I}_N \in \real^{N \times N}$.
The \mx{0} and \mx{1} denote matrices (or vectors) with all entries of zeros and ones, respectively, where the dimensions of the matrices are either known from the context or indicated by an additional subscript.

The superscript $\Tpose$ denotes the matrix transpose. 
The element-wise (Hadamard) product, element-wise division, and Kronecker products are denoted by $\mx{A} \hadamard \mx{B}$, $\mx{A} \hadamardd \mx{B}$, and $\mx{A} \kron \mx{B}$, respectively.
We use $\vectorize{\mx{A}} \in \real^{MN} = \sum_{n=1}^N \vect{e}_n \kron \left(\mx{A} \vect{e}_n\right)$, where $\vect{e}_n \in \real^{N}$, to compactly denote the the stacking of the columns of $\mx{A} \in \real^{M \times N}$.
Additionally, if a subset of the rows in \mx{A} are listed in the vector $\vectgk{\alpha} = \left[\alpha_1, \ldots, \alpha_d \right]^\Tpose \in \nat^{d}$, where $0 \leq d \leq M$ with a corresponding indexing matrix $\mx{E}_{\left[\vectgk{\alpha} \right]} = \left[\vect{e}_{\alpha_1}, \ldots, \vect{e}_{\alpha_{d}}\right]^\Tpose \in \real^{d \times M}$, then $\mx{E}_{\left[\vectgk{\alpha} \right]} \mx{A}$ selects the subset of rows in \mx{A} indicated by \vectgk{\alpha}.  
For compactness, we use $\vectorizesub{\mx{A}}{\vectgk{\alpha}} \triangleq \vectorize{\mx{E}_{\left[\vectgk{\alpha} \right]} \mx{A}}$.
The complementing subset of $\vectgk{\alpha}$ is indicated by $\vectgk{\alpha}^c \in \nat^{M-d}$.
 A diagonal matrix with entries given by \vect{d} is denoted by $\Diag{\vect{d}} = \sum_{n=1}^N \vect{e}_n \vect{e}_n^\Tpose \vect{d} \vect{e}_n^\Tpose$. 
The square matrix, \mx{A}, has diagonal entries, $\diag{\mx{A}}=\sum_{n=1}^N \vect{e}_n \vect{e}_n^\Tpose \mx{A} \vect{e}_n$, a trace, $\trace{\mx{A}}=\sum_{n=1}^N \vectelement{\diag{\mx{A}}}{n}$, and a determinant, $\det\left(\mx{A}\right)$. 
We indicate $\mx{A} - \mx{B}$  is positive definite using $\mx{A} \succ \mx{B}$ and positive semidefinite with $\mx{A} \succeq \mx{B}$. 
The operator $\abs{\cdot}$ denotes the magnitude.

For a matrix $\mx{A}$ with block structure, the matrix $\mx{A}_{m,n}$ is the $m$th row and $n$th column in the block representation of the matrix $\mx{A}$ using $M$ row partitions and $N$ column partitions.  
The special block diagonal matrix is necessarily a square matrix (implying $M=N$) that has off-diagonal partitions being zero, i.e., $\mx{A}_{m,n} = \mx{0}$ for $1\leq m\neq n \leq M$, and is denoted with the \emph{direct sum} notation, $\mx{A} = \mx{A}_{1,1} \oplus \mx{A}_{2,2} \oplus \ldots \oplus \mathbf{A}_{M,M} = \oplus\sum_{m=1}^M\mathbf{A}_{m,m}$, \cite{Horn_MatrixAnalysis_1985}. 

The common functions of random variables such as the expectation operator, entropy, and mutual information are denoted using $\expect{\cdot}$, $\entropy{\cdot}$, and $\mutinfo{\cdot}$, respectively.
A random vector $\vect{x}$ following the normal distribution with mean $\vectgk{\mu}$ and covariance matrix $\mxgk{\Sigma}$ is denoted $\vect{x} \sim \pdfnorm{\vectgk{\mu}}{\mxgk{\Sigma}}$.
We use $\vect{x} \ci \vect{y}$ to denote that a random vector $\vect{x}$ is independent of $\vect{y}$.
We use standard elementary functions such as $\log\left(\cdot\right)$, $\naturale{\cdot}$, $\Gamma\left(\cdot\right)$ 
for the natural logarithm, the anti-logarithm, and the complete Gamma function.

\section{\glsentrytext{iva} Problem Formulation}
We begin by formulating the particular \gls{jbss} framework of interest, namely  \gls{iva}, in a more general manner than previously done \cite{Kim_IVA_2007, Anderson_IVAG_2012, Hotelling_CCA_1936, XiLinLi_JointDiagGradient_2010, XiLinLi_JointDiagGradient_2010, Kim_IVA_2006, Phlypo_CDA_2013}.  The generalization allows analysis of \gls{iva} when the samples are not \gls{iid}, or alternatively when sample dependence is taken into account.

There are $K$ datasets, each containing $V$ samples, formed from the linear mixture of $N$ independent sources,
\begin{align*}
	\mx{X}^{[k]} = \mx{A}^{[k]}\mx{S}^{[k]} \in \real^{N \times V},  1\le k\le K.
\end{align*}
The entry in $n$th row and $v$th column of $\mx{S}^{[k]}$ is $s_n^{[k]}\left(v\right)$, the $n$th row of $\mx{S}^{[k]}$ is denoted with the column vector $\vect{s}_n^{[k]} = \left[s^{[k]}_n\left(1\right), \ldots, s^{[k]}_n\left(V\right)\right]^\Tpose \in \real^{V}$, and the $v$th column of $\mx{S}^{[k]}$ is denoted by the column vector $\vect{s}^{[k]}\left(v\right) = \left[s^{[k]}_1\left(v\right), \ldots, s^{[k]}_N\left(v\right)\right]^\Tpose \in \real^{N}$. 
The source matrices in each dataset can be concatenated to form $\mx{S}=\left[\left(\mx{S}^{[1]}\right)^\Tpose, \ldots, \left(\mx{S}^{[K]}\right)^\Tpose\right]^\Tpose \in \real^{NK \times V}$.
Using this notation, we can denote the \gls{jbss} data model with a single equation, namely $\mx{X} = \mx{A}\mx{S}$, where $\mx{A} = \oplus\sum_{k=1}^K\mx{A}^{[k]}$.
The invertible mixing matrices, $\mx{A}^{[k]} \in \real^{N\times N}$, and the sources $\mx{S}$ are unknown real-valued quantities to be estimated.
The $n$th \gls{scm}, $\mx{S}_{n}=\left[\vect{s}_{n}^{[1]}, \ldots, \vect{s}_{n}^{[K]}\right]^\Tpose \in \real^{K \times V}$, is independent of all other \glspl{scm}. 
Then the \gls{pdf} of the concatenated source vector, $\mx{S}$, can be written as $p\left(\mx{S}\right) = \prod_{n=1}^{N}p_n\left(\vect{S}_{n}\right)$.

The \gls{iva} solution finds $K$ demixing matrices and the corresponding source estimates for each dataset, with the $k$th ones denoted as $\mx{W}^{[k]}$ and $\mx{Y}^{[k]} \triangleq \mx{W}^{[k]} \mx{X}^{[k]}$, respectively. 
The estimate of the $n$th component from the $v$th sample of the $k$th dataset is given by $y_{n}^{[k]}\left(v\right) = \left(\vect{w}_{n}^{[k]}\right)^\Tpose \vect{x}^{[k]}\left(v\right) = \sum_{l=1}^{N}w_{n,l}^{[k]}x_{l}^{[k]}\left(v\right)$, where $\left(\vect{w}_{n}^{[k]}\right)^{\Tpose}$ is the $n$th row of $\mx{W}^{[k]}$.
Furthermore, it is  assumed that the mixing matrices possess no known relationship.

\section{\glsentrytext{iva} Objective Function}
Just as in \gls{ica}, the \gls{iva} objective function can be specified to be the maximization of the natural logarithm of the likelihood. 
Since \mx{A} is block diagonal, the estimate of the $\hat{\mx{A}}^{-1} = \mx{W} = \oplus\sum_{k=1}^K\mx{W}^{[k]}$ is block diagonal and thus we choose in the sequel to use $\mxgk{\mathcal{W}} \in \real^{N \times N \times K}$, i.e., a three-dimensional `matrix', to denote the set of parameters to be estimated.
We then have that
\begin{align}
\mathcal{L}\left(\mxgk{\mathcal{W}}\right)
&\triangleq 
\log \left(p_{\mx{X}}\left(\mx{X}\right) \right)
\nonumber
\\
& =
\log \left( \prod_{n=1}^N p_n\left(\mx{Y}_n\right) \abs{\det \mx{W}}^V \right)
\nonumber
\\
& =
\sum_{n=1}^N \log \left(  p_n\left(\mx{Y}_n\right)\right)
+ V \sum_{k=1}^K \log\abs{\det \mx{W}^{\left[k\right]}}
\label{eq:JBSS_Likelihood}
,
\end{align}
where $p_n\left(\cdot\right)$ is the model for the distribution characterizing the multivariate source $\mx{S}_n$.
Note that if $\mx{X} = \mx{A}\mx{S}$, then $\vectorize{\mx{S}} = \left(\mx{I}_V \kron \mx{A}^{-1}\right) \vectorize{\mx{X}}$, which implies $p_{\mx{X}}\left(\mx{X}; \mx{A}\right) 
= 
\abs{\det \left(\mx{I}_V \kron \mx{A}^{-1}\right)} p_{\mx{S}}\left(\left(\mx{I}_V \kron \mx{A}^{-1}\right) \vectorize{\mx{X}}\right) 
= 
\abs{\det \mx{A}^{-1}}^V p_{\mx{S}}\left( \mx{S}\right) $.

If we consider the case when $V \rightarrow \infty$, then we can define the \gls{scv} $\vect{s}_n$ as a random vector process and recall the definition of entropy rate \cite[Eq 4.10]{Cover_InfoTheoryBook_2006} so that
\begin{align}
\entropyr{\vect{s}_n} 
\triangleq 
\lim_{V \rightarrow \infty} \frac{1}{V} \entropy{\vect{s}_n\left(1\right), \ldots, \vect{s}_n\left(V\right)}
=
-\lim_{V \rightarrow \infty} \frac{1}{V} \expect{\log p_n\left(\mx{S}_n\right)}
.
\end{align}
By normalizing the likelihood objective function by $V$ and considering the limit,
\begin{align}
\mathcal{C}_{\text{IVA}}\left(\mxgk{\mathcal{W}}\right) & \triangleq -\lim_{V \rightarrow \infty} \dfrac{1}{V} \mathcal{L}\left(\mxgk{\mathcal{W}}\right)
\nonumber \\ &
= 
 \sum_{n=1}^N \entropyr{\vect{y}_n}
- \sum_{k=1}^K \log\abs{\det \mx{W}^{\left[k\right]}}
\nonumber \\ &
= \sum_{n=1}^N \left(\sum_{k=1}^K \entropyr{y_n^{[k]}} - \mutinfor{\vect{y}_n} \right) -\sum_{k=1}^K \log\abs{\det \mx{W}^{\left[k\right]}}
.
\label{eq:IVA_EntropyRate}
\end{align}
we can observe that \gls{iva} minimizes the entropy rate of the estimated \glspl{scv} (subject to the regularization term).
This representation explains that the \gls{iva} objective function will equally weight the minimization of the source entropy rates and the maximization of the across dataset dependence measure provided by the mutual information rate of $\vect{y}_n$.
It is also clear that the mutual information rate portion of the \gls{iva} objective function is responsible for resolving the permutation ambiguity across multiple datasets, since without the mutual information rate of the \glspl{scv} the objective function would be identical to using \gls{ica} on each of the $K$ datasets.
This representation will be useful in our identifiability discussion in Section \ref{sec:IvaIdentification}.

In the sequel, we will use the multivariate score function 
$\mxgk{\Phi}_n 
\triangleq 
\mxgk{\Phi}_n\left(\mx{Y}_n\right) 
=
-\partial \log \left(p_{n}\left(\mx{Y}_n\right)\right) / \partial \mx{Y}_n
 \in \real^{K \times V}$ and
 $\vectgk{\phi}^{[k]}_n = \mxgk{\Phi}_n^\Tpose \vect{e}_k$.
 
\section{\glsentrytext{iva} Fisher Information Matrix}
Here we derive the \gls{fim} of \eqref{eq:JBSS_Likelihood} \gls{wrt} $\mxgk{\mathcal{W}}$.  The $KN^2$ parameters result in $KN^2 \times KN^2$ dimension \gls{fim} with the entry associated with $w^{[k_1]}_{m_1,n_1}$ and $w^{[k_2]}_{m_2,n_2}$ denoted by and computed as:
\begin{align}
\left[\mx{F}\left(\mxgk{\mathcal{W}}\right)\right]^{k_1,m_1,n_1}_{k_2,m_2,n_2} \triangleq 
\expect{ 
\frac{\partial \mathcal{L}\left(\mxgk{\mathcal{W}}\right)}{\partial w^{[k_1]}_{m_1,n_1}}
\frac{\partial \mathcal{L}\left(\mxgk{\mathcal{W}}\right)}{\partial w^{[k_2]}_{m_2,n_2}}
}
.
\label{eq:FIM_W}
\end{align}

For the purposes of determining identifiability and the performance bound, we need only consider the \gls{fim} locally around a solution, i.e., $\mx{W} = \mx{A}^{-1}$, where $\mx{A}^{-1}$ and \mx{W} are ``freely" chosen as to alleviate all scale and permutation ambiguities.
In general, this leads to a complex expression that depends on \mx{A}; fortunately this complexity is unnecessary.  
Due to the invariance of the \gls{icrlb} on $\mx{G}=\mx{W}\mx{A}$ \gls{wrt} the mixing matrix $\mx{A} = \oplus\sum_{k=1}^K \mx{A}^{[k]}$, we need only consider $\mx{A} = \mx{I}$, i.e., the \gls{crlb} of $\mx{G}$ depends only on the statistics of the sources, \cite{Cardoso_Equivariant_1996}.
Thus the matrix of interest is
\begin{align}
\left[\mx{F}\right]^{k_1,m_1,n_1}_{k_2,m_2,n_2} 
& \triangleq
\left. \left[\mx{F}\left(\mxgk{\mathcal{W}}\right)\right]^{k_1,m_1,n_1}_{k_2,m_2,n_2} 
\right|_{\mx{A} = \mx{I},\mx{W}=\mx{I}}
.
\end{align}
 
It will prove useful to define
$\mathcal{K}_{m,n}^{[k_1,k_2]} \triangleq \frac{1}{V} \expect{\left(\vectgk{\phi}_{m}^{[k_1]}\right)^\Tpose \vect{s}^{[k_1]}_{n} \left(\vect{s}^{[k_2]}_{n}\right)^\Tpose
\vectgk{\phi}_{m}^{[k_2]}} 
, \: 1 \leq m, n \leq N$, 
to describe the form of the block diagonal \gls{fim} compactly. 
In Appendix \ref{sec:DeriveFIM}, we show that the first $N$ block entries of the \gls{fim} are given by  $\mx{F}_{n}  \triangleq \covariance{\diag{\mxgk{\Phi}_n \mx{S}_n^\Tpose - \mx{I}_V}} = V \left(\mxgk{\mathcal{K}}_{n,n} -V \mx{1}_{K \times K}\right) \in \real^{K \times K}$ and the remaining block entries are defined for $n>m$ as
\begin{align}
\mx{F}_{m,n} & \triangleq 
\covariance{\left[\begin{array}{cc}
\diag{\mxgk{\Phi}_m \mx{S}_n^\Tpose}  \\
\diag{\mxgk{\Phi}_n \mx{S}_m^\Tpose}
\end{array}\right]}
=V \left[\begin{array}{cc}
\mxgk{\mathcal{K}}_{m,n} & \mx{I}_K \\
\mx{I}_K       & \mxgk{\mathcal{K}}_{n,m}
\end{array}\right]
\label{eq:IVAFIM_mn},
 \end{align}
where the $\left(k_1,k_2\right)$ entry of $\mxgk{\mathcal{K}}_{m,n} \in \real^{K \times K}$ is $V^{-1} \ \trace{\mxgk{\Gamma}_{m}^{[k_2,k_1]} \mx{R}_{n}^{[k_1,k_2]}}$ when $m \neq n$, $\mx{R}_{n}^{[k_1,k_2]} \triangleq \expect{\vect{s}_n^{[k_1]} \left(\vect{s}_n^{[k_2]}\right)^\Tpose} \in \real^{V \times V}$, and
$\mxgk{\Gamma}_{n}^{[k_1,k_2]} 
\triangleq 
\expect{\vectgk{\phi}^{[k_1]}_n \left(\vectgk{\phi}^{[k_2]}_n\right)^\Tpose}
\in \real^{V \times V}$.

The form of the \gls{fim} is a multivariate extension of the single dataset forms given in \cite{Tichavsky_CRLBforFastICA_2006, Ollila_CRLBforICA_2008, Yeredor_BSSgauss_2010, Comon_HandbookBSS_2010}.
The \gls{fim} has a form that is a block matrix version of the single dataset result, e.g., see \figref{fig:fig_FIM_N3} and compare to the similar form given in \cite{Loesch_ComplexCrlbICA_2013} for complex-valued \gls{ica}.
The $2 \times 2$ blocks with ones in the off-diagonal elements and pair-wise cross terms in the two diagonal elements of the \gls{ica} \gls{fim} are here replaced with $2 \times 2$ block matrices with identity matrices in the off-diagonal blocks and the cross terms in the two diagonal block matrices, i.e., $\mx{F}_{m,n}$.
 
\section{\glsentrytext{iva}
 Identification Conditions}
 \label{sec:IvaIdentification}
The identification of sources in (real-valued) \gls{ica} is possible so long as no two sources are Gaussian with proportional covariance matrices \cite[Chapter 4]{Comon_HandbookBSS_2010}.
When sources are said to be identifiable for \gls{ica}, this means that the sources can be recovered up to a scale factor and arbitrary ordering, i.e., the true mixing matrix $\mx{A}_0$ can be identified upto $\mx{A}_0 \mxgk{\Lambda} \mx{P}$, where $\mxgk{\Lambda}$ is any nonsingular diagonal matrix and $\mx{P}$ is any permutation matrix.

Since the the model structure of  \gls{iva} is a generalization of the model structure for \gls{ica}, we expect a generalization of the identification conditions for \gls{ica}.
Intuitively, the identification conditions for \gls{iva} are related to the dependence of the sources across the datasets.
More specifically, when sources possess dependence across datasets we expect that these estimated sources can be `aligned'---this is the original motivation of \gls{iva} \cite{Kim_IVA_2006, Hiroe_AlmostIVA_2006}.  
However, if there are sources for which no alignment exhibits dependence, then under the \gls{ica} identification conditions sources can be separated but not necessarily aligned.  
That is, without dependence across datasets the estimated sources of \gls{iva} would be no different than using \gls{ica} on each dataset individually since there is no dependency to exploit.
The identification conditions, which we present in this section, capture both cases, i.e., when there is or is not dependence between sources across datasets.

To discuss identifiability of \gls{iva}, we need to provide a notation that allows us to indicate a particular subset of rows in an \gls{scm}.  
For this section, we let $\vectgk{\alpha} = \left[\alpha_1 \ldots \alpha_{d_\alpha} \right]^\Tpose \in \nat^{K_\alpha}$, where $0 \leq K_\alpha \leq K$. 
The complementing subset of $\vectgk{\alpha}$ in $\left\{1, \ldots, K\right\}$ is indicated by $\vectgk{\alpha}^c \in \nat^{K-K_\alpha}$. 
The \gls{iva} identification conditions use the following definition:
\begin{definition}[\emph{$\vectgk{\alpha}$-Gaussian}]
A source, $\mx{S} \in \real^{K \times V}$, has an \emph{$\vectgk{\alpha}$-Gaussian} component when 
$ \vectorizesub{\mx{S}}{\vectgk{\alpha}} \ci \vectorizesub{\mx{S}}{\vectgk{\alpha}^c}$, and $\vectorizesub{\mx{S}}{\vectgk{\alpha}} \sim \pdfnorm{\vect{0}}{\mx{R}_\alpha}$, where $\mx{R}_\alpha = \expect{\vectorizesub{\mx{S}}{\vectgk{\alpha}} \vectorizesubT{\mx{S}}{\vectgk{\alpha}}} \in \real^{K_\alpha V \times K_\alpha V}$ is nonsingular.
\end{definition}
The $\vectgk{\alpha}$-Gaussian definition is used to identify that there exist a subset of rows in an \gls{scm} that is independent of the other rows in the same \gls{scm} and that the given subset follows a multivariate Gaussian distribution.
The theorem stating the \gls{iva} identification conditions and its proof follow.
\begin{theorem}[\Gls{iva} Nonidentifiability]
\label{thm:iva_id}
The sources cannot be identified \gls{iff} $\exists \: \vectgk{\alpha} \neq \emptyset$ and $\exists \: m\neq n$ such that
 $\mx{S}_m$ and $\mx{S}_n$ have \emph{$\vectgk{\alpha}$-Gaussian} components for which
 $\mx{R}_{m,\alpha} = \left(\mx{I}_V \kron \mx{D}\right) 
\mx{R}_{n,\alpha}
\left(\mx{I}_V \kron \mx{D}\right)
\in \real^{K_\alpha V \times K_\alpha V}
$,
 where $\mx{D} \in \real^{K_\alpha \times K_\alpha}$ is any full rank diagonal matrix.
\end{theorem}

\begin{IEEEproof}[Proof of IVA Nonidentifiability]
Given the \gls{fim} \eqref{eq:app:IVAFIM}, \eqref{eq:app:IVAFIM_n}, \eqref{eq:app:IVAFIM_mn}, since $\mx{F}_{m,n}$ is a covariance matrix, it must be positive semidefinite and is singular \gls{iff} $\exists \left(\vect{a}, \vect{b}\right) \neq \left(\vect{0}, \vect{0}\right): 
\vect{a}^\Tpose \diag{\mxgk{\Phi}_m \mx{S}_n^\Tpose} 
-
\vect{b}^\Tpose \diag{\mxgk{\Phi}_n \mx{S}_m^\Tpose} = 0, \: 
\forall \:
\mx{S}_m \in \Omega_{\mx{S}_m}, \mx{S}_n \in \Omega_{\mx{S}_n}$, where $\Omega_{\mx{X}}$ denotes the sample space of the random matrix $\mx{X}$. 

It is convenient to rewrite the following:
\begin{align}
\diag{\mxgk{\Phi}_m \mx{S}_n^\Tpose} 
&=
\diag{\sum_{v=1}^V \vectgk{\phi}_{m}\left(v\right) \vect{s}_{n}^\Tpose \left(v\right)}
=
\sum_{v=1}^V \vectgk{\phi}_{m}\left(v\right) \hadamard \vect{s}_{n}\left(v\right),
\end{align}
where $\vect{s}_{n}\left(v\right)$ and $\vectgk{\phi}_{m}\left(v\right)$ denote the $v$th columns of $\mx{S}_n$ and $\mxgk{\Phi}_m$, respectively.

Hence, the following statements are all equivalent conditional on $\exists \left(\vect{a}, \vect{b}\right) \neq \left(\vect{0}, \vect{0}\right):$
\begin{align}
&&&\mx{F}_{m,n} \textrm{ is singular}
\label{eq:JBSSIdProof-a1}
\\
&\Leftrightarrow&
0 &= \vect{a}^\Tpose \diag{\mxgk{\Phi}_m \mx{S}_n^\Tpose} 
-
\vect{b}^\Tpose \diag{\mxgk{\Phi}_n \mx{S}_m^\Tpose} 
\label{eq:JBSSIdProof-a}
\\
&\Leftrightarrow&
0 &=
\vect{a}^\Tpose \sum_{v=1}^V \vectgk{\phi}_{m}\left(v\right) \hadamard \vect{s}_{n}\left(v\right)
-
\vect{b}^\Tpose \sum_{q=1}^V \vectgk{\phi}_{n}\left(q\right) \hadamard \vect{s}_{m}\left(q\right)
\label{eq:JBSSIdProof-b}
\\
&\Leftrightarrow&
0 &=
\left(\vect{1}_V \kron \vect{a}\right)^\Tpose 
\left(\vectorize{\mxgk{\Phi}_m} \hadamard \vectorize{\mx{S}_n}\right)
- 
\left(\vect{1}_V \kron \vect{b}\right)^\Tpose 
\left(\vectorize{\mxgk{\Phi}_n} \hadamard \vectorize{\mx{S}_m}\right)
\label{eq:JBSSIdProof-c}
\\
&\Leftrightarrow&
0 &=
\vectorizesubT{\mxgk{\Phi}_m}{\vectgk{\alpha}}
\left(\mx{I}_V \kron \mx{D}_{\vect{a}\left[\vectgk{\alpha} \right]}\right) \vectorizesub{\mx{S}_n}{\vectgk{\alpha}}
-
\vectorizesubT{\mxgk{\Phi}_n}{\vectgk{\beta}}
\left(\mx{I}_V \kron \mx{D}_{\vect{b}\left[\vectgk{\beta} \right]}\right) \vectorizesub{\mx{S}_m}{\vectgk{\beta}}
\label{eq:JBSSIdProof-da}
\\
&\Leftrightarrow&
0 &=
\vectorizesubT{\mxgk{\Phi}_m}{\vectgk{\alpha}}
\left(\mx{I}_V \kron \mx{D}_{\vect{a}\left[\vectgk{\alpha} \right]}\right) \vectorizesub{\mx{S}_n}{\vectgk{\alpha}}
-
\vectorizesubT{\mxgk{\Phi}_n}{\vectgk{\alpha}}
\left(\mx{I}_V \kron \mx{D}_{\vect{b}\left[\vectgk{\alpha} \right]}\right) \vectorizesub{\mx{S}_m}{\vectgk{\alpha}}
\label{eq:JBSSIdProof-d}
\\
&\Leftrightarrow&
0 &=
\vectorizesubT{\mx{S}_m}{\vectgk{\alpha} } \mx{R}_{m,\alpha}^{-1}
\left(\mx{I}_V \kron \mx{D}_{\vect{a}\left[\vectgk{\alpha} \right]}\right) \vectorizesub{\mx{S}_n}{\vectgk{\alpha}}
-
\vectorizesubT{\mx{S}_n}{\vectgk{\alpha}} \mx{R}_{n,\alpha}^{-1}
\left(\mx{I}_V \kron \mx{D}_{\vect{b}\left[\vectgk{\alpha} \right]}\right) \vectorizesub{\mx{S}_m}{\vectgk{\alpha}} 
\label{eq:JBSSIdProof-e}
\\
&\Leftrightarrow&
0 &=
 \mx{R}_{m,\alpha}^{-1}
\left(\mx{I}_V \kron \mx{D}_{\vect{a}\left[\vectgk{\alpha} \right]}\right) 
- 
\left(\mx{I}_V \kron \mx{D}_{\vect{b}\left[\vectgk{\alpha} \right]}\right)
\mx{R}_{n,\alpha}^{-1}
\label{eq:JBSSIdProof-f}
\\
&\Leftrightarrow&
 \mx{R}_{m,\alpha} &=
\left(\mx{I}_V \kron \mx{D}_{\vect{a}\left[\vectgk{\alpha} \right]}\right) 
\mx{R}_{n,\alpha}
\left(\mx{I}_V \kron \mx{D}^{-1}_{\vect{b}\left[\vectgk{\alpha} \right]}\right)
\label{eq:JBSSIdProof-g}
\\
&\Leftrightarrow&
 \mx{R}_{m,\alpha} &= 
\left(\mx{I}_V \kron \mx{D}\right) 
\mx{R}_{n,\alpha}
\left(\mx{I}_V \kron \mx{D}\right)
\label{eq:JBSSIdProof-h}
,
\end{align}
where
$\mx{D}_{\vect{a}\left[\vectgk{\alpha} \right]} \triangleq \Diag{\vect{a}\left[\vectgk{\alpha} \right]}$, $\mx{D}_{\vect{b}\left[\vectgk{\alpha} \right]} \triangleq \Diag{\vect{b}\left[\vectgk{\alpha} \right]}$,
$\vectgk{\alpha} \in \nat^{K_\alpha}$, and $\vectgk{\beta}\in \nat^{K_\beta}$.

It is straightforward to observe that \eqref{eq:JBSSIdProof-a1}, \eqref{eq:JBSSIdProof-a}, \eqref{eq:JBSSIdProof-b}, and \eqref{eq:JBSSIdProof-c} are equivalent expressions.  
From the relationship $\left(\vect{x} \kron \vect{y}\right)^\Tpose \left(\vect{w} \hadamard \vect{z}\right) = \vect{w}^\Tpose \left(\Diag{\vect{x}} \kron \Diag{\vect{y}} \right) \vect{z}$, the expression in  \eqref{eq:JBSSIdProof-da} holds only when $\vectgk{\alpha} = \vectgk{\beta}$, i.e., the zero entries of $\vect{a}$ and $\vect{b}$ are at the same locations.  
See Lemma \ref{lem:jbss} below to explain \eqref{eq:JBSSIdProof-e}.  
Since \eqref{eq:JBSSIdProof-e} must hold for all possible values of $\vectorizesub{\mx{S}_m}{\vectgk{\alpha}}$ and $\vectorizesub{\mx{S}_n}{\vectgk{\alpha}}$, \eqref{eq:JBSSIdProof-f} must hold.  
Equation \eqref{eq:JBSSIdProof-g} is equivalent since all entries of $\vect{b}\left[\vectgk{\alpha}\right]$ are nonzero by \eqref{eq:JBSSIdProof-d}.  
Lastly, since  $\mx{R}_{m,\alpha}$ is symmetric we must have that either $\mx{R}_{n,\alpha}$ is diagonal or $\mx{D}_{\vect{a}\left[\vectgk{\alpha} \right]} = 
\left(\mx{D}_{\vect{b}\left[\vectgk{\alpha} \right]}\right)^{-1}$.  In either case \eqref{eq:JBSSIdProof-h} holds.
\end{IEEEproof}

\begin{lemma}
\label{lem:jbss}
For $m \neq n$,  
\begin{align}
\vectorizesubT{\mxgk{\Phi}_m}{\vectgk{\alpha}}
\left(\mx{I}_V \kron \mx{D}_{\vect{a}\left[\vectgk{\alpha} \right]}\right) \vectorizesub{\mx{S}_n}{\vectgk{\alpha}} 
&=
\vectorizesubT{\mxgk{\Phi}_n}{\vectgk{\alpha}}
\left(\mx{I}_V \kron \mx{D}_{\vect{b}\left[\vectgk{\alpha} \right]}\right) \vectorizesub{\mx{S}_m}{\vectgk{\alpha}}
\label{eq:lem_jbss_1}
\end{align}
holds \gls{iff}
\begin{align}
\vectorizesubT{\mx{S}_m}{\vectgk{\alpha}} \mx{R}_{m,\alpha}^{-1}
\left(\mx{I}_V \kron \mx{D}_{\vect{a}\left[\vectgk{\alpha} \right]}\right) \vectorizesub{\mx{S}_n}{\vectgk{\alpha}}
&=
\vectorizesubT{\mx{S}_n}{\vectgk{\alpha}} \mx{R}_{n,\alpha}^{-1}
\left(\mx{I}_V \kron \mx{D}_{\vect{b}\left[\vectgk{\alpha} \right]}\right) \vectorizesub{\mx{S}_m}{\vectgk{\alpha}} 
\label{eq:lem_jbss_2}
\end{align}
and $\mx{S}_m$ and $\mx{S}_n$ each have an \emph{$\vectgk{\alpha}$-Gaussian} component.  
\end{lemma}
\begin{IEEEproof}
$\left(\Rightarrow\right)$ Since the left-hand side of \eqref{eq:lem_jbss_1} is linear in $\vectorizesub{\mx{S}_n}{\vectgk{\alpha}} $ we must have that $\vectorizesub{\mxgk{\Phi}_n}{\vectgk{\alpha}}$ is not a function of $\vectorizesub{\mx{S}_n}{\vectgk{\alpha}^c} $ and it is necessarily linear in $\vectorizesub{\mx{S}_n}{\vectgk{\alpha}} $, i.e., 
$\mx{S}_n$ has \emph{$\vectgk{\alpha}$-Gaussian} component.  
By symmetry, the same can be concluded about $\mx{S}_m$.

$\left(\Leftarrow\right)$ If $\mx{S}_n$ has \emph{$\vectgk{\alpha}$-Gaussian} component then $\vectorizesub{\mxgk{\Phi}_n}{\vectgk{\alpha}}  = \mx{R}_n^{-1} \vectorizesub{\mx{S}_n}{\vectgk{\alpha}}$.
\end{IEEEproof}

It is noteworthy to mention that the \gls{iva} identification conditions admit sources for which the distribution can be factored, i.e., $p_n\left(\mx{S}_n\right) = \prod_{q=1}^{
Q} p_{n_q}\left(\vectorizesub{\mx{S}_n}{\mathcal{Q}_q}\right)$, where $\left\{\mathcal{Q}_1, \mathcal{Q}_2, \ldots, \mathcal{Q}_Q \right\}$, $\mathcal{Q}_q \subset \left\{1, ..., K\right\}$, $\mathcal{Q}_q \cap \mathcal{Q}_{q'} = \emptyset \: \forall q \neq q'$, and $\cup_{q=1}^Q \mathcal{Q}_q = \left\{1, ..., K\right\}$.  If, for example $Q=K$, then \gls{iva} would produce the same identification conditions as \gls{ica} on each dataset individually.  
Stated differently, identifiability of \gls{iva} does not require the sources to possess dependence across datasets.

Recalling that a prime motivation for considering the \gls{iva} formulation is to determine when the sources can be aligned in a common way across all datasets, i.e., under what conditions is $\hat{\mx{A}}^{[k]} = \left(\mx{W}^{[k]}\right)^{-1} = \mx{A}^{[k]} \mx{P} \mxgk{\Lambda}^{[k]}$, where $\mxgk{\Lambda}^{[k]}$ is any full rank diagonal matrix and \mx{P} is a permutation matrix commonly shared by all datasets.
The common permutation identification condition is given in the next theorem which uses the following definition:
\begin{definition}[\emph{$\vectgk{\alpha}$-independent}]
A source, $\mx{S} \in \real^{K \times V}$, is \emph{$\vectgk{\alpha}$-independent}  when 
$ \vectorizesub{\mx{S}}{\vectgk{\alpha}} \ci \vectorizesub{\mx{S}}{\vectgk{\alpha}^c}$.
\end{definition}
The $\vectgk{\alpha}$-independent definition is used to identify that there exist a subset of rows in an \gls{scm} (or \gls{scv}) that is independent of the other rows in the \gls{scm} (\gls{scv}).
\begin{theorem}[Common Permutation Matrix for \Gls{iva}]
\label{thm:iva_permutation}
Assuming the \gls{iva} identification conditions of Theorem \ref{thm:iva_id} are satisfied, i.e., in the limit as $V \rightarrow \infty$ so that $\left(\mx{W}^{[k]}\right)^{-1} = \mx{A}^{[k]} \mx{P}^{[k]} \mxgk{\Lambda}^{[k]}$:\\
The permutation matrix associated with each dataset is common \gls{iff} $\forall \ m \neq n \ \nexists \ \vectgk{\alpha} \neq \emptyset$ such that both $\vect{s}_m$ and $\vect{s}_n$ are \emph{$\vectgk{\alpha}$-independent}.
\end{theorem}
\begin{IEEEproof}
The objective function given in \eqref{eq:IVA_EntropyRate} makes it clear that any permutation matrix at most effects the  \mutinfor{\vect{y}_n} term.
Furthermore, we only need consider permutation matrices that can achieve the global minimum.
The proof is by contradiction (in both directions):
\begin{align}
&& \exists 1 \leq k_1 \neq k_2 \leq K &: \mx{P}^{[k_1]} \neq \mx{P}^{[k_2]}
\\
&\Leftrightarrow& 
\mutinfor{\vect{s}_m^{[\vectgk{\alpha}]}; \vect{s}_n^{[\vectgk{\alpha}^c]}} + \mutinfor{\vect{s}_n^{[\vectgk{\alpha}]}; \vect{s}_m^{[\vectgk{\alpha}^c]}} 
&= 
\mutinfor{\vect{s}_m^{[\vectgk{\alpha}]}; \vect{s}_m^{[\vectgk{\alpha}^c]}} + \mutinfor{\vect{s}_n^{[\vectgk{\alpha}]}; \vect{s}_n^{[\vectgk{\alpha}^c]}} 
\\
&\Leftrightarrow &
0 
&=
\mutinfor{\vect{s}_m^{[\vectgk{\alpha}]}; \vect{s}_m^{[\vectgk{\alpha}^c]}} + \mutinfor{\vect{s}_n^{[\vectgk{\alpha}]}; \vect{s}_n^{[\vectgk{\alpha}^c]}} 
\\
&\Leftrightarrow& 
\mx{S}_m \text{ and } \mx{S}_n & \text{ are $\vectgk{\alpha}$-independent}
\end{align}
We have used the fact that $\mutinfor{\mathcal{X};\mathcal{Y}} \geq 0$ with equality \gls{iff} $\mathcal{X} \ci \mathcal{Y}$, which implies by the assumption of \gls{iva} that $\mutinfor{\vect{s}_i^{[\vectgk{\alpha}_1]}; \vect{s}_j^{[\vectgk{\alpha}_2]}} = 0 \ \forall i \neq j, \vectgk{\alpha}_1, \ \vectgk{\alpha}_2 $, where $\vectgk{\alpha}_1$ and $\vectgk{\alpha}_2$ are any indexing sets.
\end{IEEEproof}
Thus, Theorem \ref{thm:iva_permutation} provides an additional restriction on the sources (in a pairwise manner) which is required when the estimated dependent sources across all datasets are to be `aligned'.

\subsection{Special Cases}
It is now insightful to consider important special cases of \gls{iva} with regard to the identification conditions.
We begin by considering the case when the $V$ samples are \gls{iid}.  This is equivalent to having $V=1$, which implies that the identification conditions can be derived as a special case of Theorem \ref{thm:iva_id}.
\begin{theorem}[\Gls{iva} Nonidentifiability with \gls{iid} Samples]
\label{thm:ivaiid_id}
The sources cannot be identified \gls{iff} $\exists \: \vectgk{\alpha} \neq \emptyset$ and $\exists \: m\neq n$ such that
 $\vect{s}_m$ and $\vect{s}_n$ have \emph{$\vectgk{\alpha}$-Gaussian} components and 
 $\mx{R}_{m,\alpha} = \mx{D} 
\mx{R}_{n,\alpha}
\mx{D}
 \in \real^{K_\alpha \times K_\alpha}$,
 where $\mx{D}$ is any full rank diagonal matrix.
\end{theorem}

Another special case of interest is when $K=1$, yielding the same formulation as \gls{ica} assuming sample-to-sample dependence, i.e., not \gls{iid} samples, the most general form for real-valued \gls{ica}.  

\begin{theorem}[\Gls{ica} Nonidentifiability \cite{Comon_HandbookBSS_2010}, \cite{Afsari_JointDiag_2008}]
\label{thm:ica_id}
The sources cannot be identified \gls{iff} $\exists \: m\neq n$ such that
 $\vect{s}_m \in \real^V$ and $\vect{s}_n \in \real^V$ are \emph{Gaussian} and
 $\mx{R}_m = \delta^2 
\mx{R}_n
\in \real^{V \times V}$,
 where $\delta \neq 0$.
\end{theorem}
It can be verified that the identification conditions of Theorem \ref{thm:ica_id} are consistent with the results found in \cite[Chapter 4]{Comon_HandbookBSS_2010} and \cite{Afsari_JointDiag_2008}.

Another special case of interest is when $K=1$, and assuming \gls{iid} samples.
\begin{theorem}[\Gls{ica} Nonidentifiability with \gls{iid} Samples \cite{Comon_ICAConcept_1994}]
\label{thm:ica_id_iid}
The sources cannot be identified \gls{iff} $\exists \: m\neq n$ such that
 $s_m \in \real$ and $s_n \in \real$ are Gaussian.
\end{theorem}
The claim of Theorem \ref{thm:ica_id_iid}, originally given in \cite{Comon_ICAConcept_1994}, states the well known result for \gls{ica} that at most one source can be Gaussian for identification of all \gls{iid} sources.
Algorithms based on the \gls{iid} assumption using higher-order statistics have been the most widely exploited type of diversity in the derivation of \gls{ica} algorithms.

Additional \emph{diversity} can extend the \gls{iva} and \gls{ica} identification conditions. An example is when data is complex-valued, a case we do not consider in this paper.

\section{\glsentrytext{crlb} and \glsentrytext{icrlb}}
\label{sec:CRLB}
The \gls{crlb} associated with the parameter vector \mxgk{\Theta} is the inverse of the \gls{fim}, i.e., $\covariance{\hat{\mxgk{\Theta}}} \geq \mx{F}^{-1}$, where $\hat{\mxgk{\Theta}}$ is an estimator for \mxgk{\Theta}.
Due to the block diagonal structure of \eqref{eq:app:IVAFIM} we have that the inverse (if it exists, see identifiability discussion in Section \ref{sec:IvaIdentification}) of the portion of the \gls{fim} associated with the $m$th and $n$th source denoted by $\mx{F}_{m,n}$ in \eqref{eq:IVAFIM_mn} is
\begin{align*}
\mx{F}_{m,n}^{-1} = \frac{1}{V} \left[\begin{array}{cc}
\left(\mxgk{\mathcal{K}}_{m,n} - \mxgk{\mathcal{K}}_{n,m}^{-1}\right)^{-1} & * \\
*       & \left(\mxgk{\mathcal{K}}_{n,m} - \mxgk{\mathcal{K}}_{m,n}^{-1}\right)^{-1}
\end{array}\right]
. 
\end{align*}

It yields the following \gls{crlb} on the estimates of the demixing matrix quantities,
\begin{align*}
\variance{w_{m,n}^{[k]}} & \geq \frac{1}{V} \mathbf{e}_k^\Tpose \left(\mxgk{\mathcal{K}}_{m,n} - \mxgk{\mathcal{K}}_{n,m}^{-1}\right)^{-1} \mathbf{e}_k, \ 1 \leq m \neq n \leq N.
\end{align*}

For this \gls{jbss} formulation, the definition of the \gls{isr} is the same as in \gls{bss} \cite{Yeredor_BSSgauss_2010,Comon_HandbookBSS_2010}, namely:
\begin{align}
\text{ISR}_{m,n}^{[k]} & \triangleq \expect{\left(g_{m,n}^{[k]}\right)^2} \frac{\expect{\abs{\vect{s}_n^{[k]}}^2}}{\expect{\abs{\vect{s}_m^{[k]}}^2}}, \ 1 \leq m \neq n \leq N,
\end{align}
where $g_{m,n}^{[k]} = \vect{e}_m^\Tpose \mx{G}^{[k]} \vect{e}_n$ and $\mx{G}^{[k]} \triangleq \mx{W}^{[k]} \mx{A}^{[k]}$ is called the $k$th global demixing-mixing matrix.

The \gls{icrlb} for \gls{isr} is then:
\begin{align}
\text{ISR}_{m,n}^{[k]} \geq
\frac{1}{V} \mathbf{e}_k^\Tpose \left(\mxgk{\mathcal{K}}_{m,n} - \mxgk{\mathcal{K}}_{n,m}^{-1}\right)^{-1} \mathbf{e}_k \frac{\expect{\abs{\vect{s}_n^{[k]}}^2}}{\expect{\abs{\vect{s}_m^{[k]}}^2}}. 
\end{align}

Since the \emph{sources} are (potentially) multivariate in the \gls{iva} formulation, it makes sense to define the \gls{isr} according to
\begin{align*}
\text{ISR}_{m,n} \triangleq \sum_{k=1}^K \text{ISR}_{m,n}^{[k]}, \ 1 \leq m \neq n \leq N.
\end{align*}

After some simple manipulation, the following compact form for the \gls{icrlb} results:
\begin{align*}
\text{ISR}_{m,n} \geq \frac{1}{V} 
\trace{
\left(\mxgk{\mathcal{K}}_{m,n} - \mxgk{\mathcal{K}}_{n,m}^{-1}\right)^{-1}
\hadamard 
\mx{C}_n
\hadamardd
\mx{C}_m}
,
\end{align*}
where $\mx{C}_n \triangleq \expect{\mx{S}_n \mx{S}_n^\Tpose} \in \real^{K \times K}$.  
In what follows, for notational simplicity and without loss of generality, we assume the sources have equal energy within each dataset, i.e.,  $\diag{\mx{C}_n} = \diag{\mx{C}_m} \ \forall \ 1 \leq m,n \leq N$.

When the samples are \gls{iid}, then the \gls{iva} \gls{icrlb} simplifies further if we note that:  
\begin{align}
\mx{R}_{n}^{[k_1,k_2]} = \expect{\vect{s}_n^{[k_1]} \left(\vect{s}_n^{[k_2]}\right)^\Tpose} 
=
 \sigma^{[k_1,k_2]}_n \mx{I}_{V},
\end{align}
\begin{align}
\mxgk{\Gamma}_{m}^{[k_1,k_2]} 
  = 
  \expect{\vectgk{\phi}^{[k_1]}_m \left(\vectgk{\phi}^{[k_2]}_m\right)^\Tpose}
=
 \gamma^{[k_1,k_2]}_m \mx{I}_{V},
\end{align}
and for $1\leq m \neq n \leq N$,
\begin{align}
   \mathcal{K}_{m,n}^{[k_1,k_2]} 
=
\frac{1}{V}
\trace{\mxgk{\Gamma}_{m}^{[k_2,k_1]} \mx{R}_{n}^{[k_1,k_2]}}
= 
\gamma^{[k_1,k_2]}_m \sigma^{[k_1,k_2]}_n 
,
\end{align}
where $ \sigma^{[k_1,k_2]}_n \triangleq \expect{s_n^{[k_1]}\left(v\right) s_n^{[k_2]}\left(v\right)} \in \real$ and
$\gamma^{[k_1,k_2]}_m \triangleq \expect{\phi^{[k_1]}_{m}\left(v\right) \phi^{[k_2]}_{m}\left(v\right)} \in \real$ are not dependent on $v$ due to the \gls{iid} assumption. 

For the \gls{iid} \gls{iva} discussion we simplify by replacing the \gls{scm} notation with \gls{scv} notation, i.e., we define the \gls{scv}, $\vect{s}_n$, as a random vector with $V$ realizations denoted by $\vect{s}_n \left(v\right) \in \real^K$.  
In addition, the multivariate score function is denoted by $\vectgk{\phi}_m\left(\vect{s}_m\right) \in \real^{K}$.  
For now, let $\mx{R}_n = \expect{\vect{s}_n \vect{s}_n^\Tpose} \in \real^{K \times K}$ 
and 
$\mxgk{\Gamma}_m = \expect{\vectgk{\phi}_m\left(\vect{s}_m\right) \vectgk{\phi}_m^\Tpose \left(\vect{s}_m\right)} \in \real^{K \times K}$, 
from which we observe that $\mxgk{\mathcal{K}}_{m,n} = \mxgk{\Gamma}_m \hadamard \mx{R}_n = \variance{\vectgk{\phi}_m\left(\vect{s}_m\right) \hadamard \vect{s}_n}$.

The above gives the following \gls{icrlb} on the estimates of the demixing matrix entries when the samples are \gls{iid},
\begin{align*}
\text{ISR}_{m,n} \geq \frac{1}{V} 
\trace{
\left(\mxgk{\Gamma}_m \hadamard \mx{R}_n - \left(\mxgk{\Gamma}_n \hadamard \mx{R}_m\right)^{-1}\right)^{-1}
}
. 
\end{align*} 

The relationship between $\mxgk{\Gamma}$ and $\mx{R}$  given in the following lemma is the multivariate extension of the result given by \cite[Lemma 1b of Appendix B]{Ollila_CRLBforICA_2008}, which has also been given in \cite[Chapter 4]{Comon_HandbookBSS_2010}.
\begin{lemma} 
\label{lem:GammaVsInvCov} $\mxgk{\Gamma} \succeq \mx{R}^{-1}$, with equality \gls{iff} $\vectgk{\phi} = \mx{R}^{-1} \vect{s}$, i.e., $\vect{s}$ follows the Gaussian distribution.
\end{lemma}
\begin{IEEEproof}
The proof applies the extension of the Cauchy-Schwarz inequality for covariance matrices as given in \cite{Lavergne_CauchySchwarzIneq_2008}.  
Specifically, $
\mxgk{\Gamma} - \expect{\vectgk{\phi} \vect{s}^\Tpose} \mx{R}^{-1} \expect{\vect{s} \vectgk{\phi}^\Tpose} \succeq \mx{0}$, with equality \gls{iff} $\vectgk{\phi} = \expect{\vectgk{\phi} \vect{s}^\Tpose} \mx{R}^{-1} \mx{s}$.  
By noting that $\expect{\vect{s} \vectgk{\phi}^\Tpose} = \mx{I}$ we arrive at the assertion.
\end{IEEEproof}
From this lemma, we see that a measure of non-Gaussianity (or higher-order statistics) is captured by the `difference' between $\mxgk{\Gamma}$ and $\mx{R}^{-1}$.  
Next, we show for elliptical distributions---a broad class of source distributions---how this non-Gaussianity measure can be captured by a scalar quantity.

The \gls{pdf} (assuming it exists) for a zero-mean random vector following the elliptical distribution is
\begin{align}
p\left(\vect{x}\right)
&=
\frac{c_K}{
\sqrt{\det{\mxgk{\Sigma}}}}
h_e\left(
\vect{x}^\Tpose 
\mxgk{\Sigma}^{-1} 
\vect{x}
\right),
\label{eq:ellipticalpdf}
\end{align}
where  $\mxgk{\Sigma} \in \real^{K \times K}$ is the positive definite matrix frequently termed the dispersion matrix, $h_e$ is some nonnegative function, and $c_K$ denotes the constant that makes \eqref{eq:ellipticalpdf} integrate to one.
If the covariance matrix, $\expect{\vect{x} \vect{x}^\Tpose} = \mx{R}$, exists, then for any elliptical 
distribution it is a scalar multiple of the dispersion matrix, i.e., $\mx{R} = \rho \mxgk{\Sigma}$, where $\rho > 0$.
Then the score function, $\vectgk{\phi}\left(\vect{x}\right) \triangleq - \partial \log p\left(\vect{x}\right) / \partial \vect{x} =
g\left(\vect{x}^\Tpose \mxgk{\Sigma}^{-1} \vect{x} \right) \mxgk{\Sigma}^{-1} \vect{x}$,
where $g\left(u \right) =- 2 \tfrac{1}{ h_e\left(u\right)}\tfrac{d h_e\left(u\right)}{d u}$.

For elliptical distributions (see Appendix \ref{sec:EllipticalGammaMx}), $
 \mxgk{\Gamma}
 = 
 \kappa \mx{R}^{-1}, K \geq 2$,
where $\kappa \triangleq  \expect{ g^2(r^2) r^{K+1}} 
\frac{2\pi^{K/2}}{K
\Gamma\left(K/2\right)} \rho$.
By application of Lemma \ref{lem:GammaVsInvCov} this implies that $\kappa \geq 1$ with equality \gls{iff} Gaussian\footnote{
Under the Gaussian \gls{scv} data-model assumption, $\expect{\vectgk{\phi} \vectgk{\phi}^\Tpose }
 = 
 \expect{\mx{R}^{-1} \vect{s} \vect{s}^\Tpose \mx{R}^{-1}} 
 = 
 \mx{R}^{-1}$.}.
Therefore, the \gls{icrlb} for \gls{isr} with elliptical sources is
\begin{align*}
\text{ISR}_{m,n} \geq \frac{1}{V} 
\trace{
\left(\kappa_m \mx{R}_m^{-1} \hadamard \mx{R}_n - \left(\kappa_n \mx{R}_n^{-1} \hadamard \mx{R}_m\right)^{-1}\right)^{-1}}
.
\end{align*}
For this performance bound we provide the following theorem.
\begin{theorem}
\label{thm:ISRbound}
If two \glspl{scv} follow distributions from the elliptical family with covariance matrices, $\mx{R}_m$ and $\mx{R}_n$, then $\text{ISR}_{m,n}$ is less than or equal to the $\text{ISR}_{m,n}$ associated with Gaussian \glspl{scv} having the same covariance matrices.
\end{theorem}
\begin{IEEEproof}
See \cite{Anderson_IVAG_2012} for proof that $\mx{R}_m^{-1} \hadamard \mx{R}_{n} - \left( \mx{R}_n^{-1} \hadamard \mx{R}_{m}\right)^{-1} \succeq 0$.  
For elliptically distributed sources, via Lemma \ref{lem:GammaVsInvCov}, we have that $\kappa_m \geq 1$ and $\kappa_n \geq 1$, thus 
\begin{align*}
\kappa_
m \mx{R}_m^{-1} \hadamard \mx{R}_{n} - \kappa_n^{-1} \left( \mx{R}_n^{-1} \hadamard \mx{R}_{m}\right)^{-1} 
&\succeq
\mx{R}_m^{-1} \hadamard \mx{R}_{n} - \left( \mx{R}_n^{-1} \hadamard \mx{R}_{m}\right)^{-1}
\\
\left(\kappa_m \mx{R}_m^{-1} \hadamard \mx{R}_{n} - \kappa_n^{-1} \left( \mx{R}_n^{-1} \hadamard \mx{R}_{m}\right)^{-1}\right)^{-1} &\preceq 
\left(\mx{R}_m^{-1} \hadamard \mx{R}_{n} - \left( \mx{R}_n^{-1} \hadamard \mx{R}_{m}\right)^{-1}\right)^{-1},
\end{align*}
and since $\mx{A} \preceq \mx{B}$, it implies $\vect{x}^\Tpose \mx{A} \vect{x} \leq \vect{x}^\Tpose \mx{B} \vect{x}, \forall \vect{x}$, and thus $\trace{\mx{A}} \leq \trace{\mx{
B}}$.
\end{IEEEproof}
A special case, which arrives at a form directly analogous to the \gls{ica} form, occurs when $\mx{R}_m = \mx{R}_n = \mx{I}_K$:
\begin{align}
\text{ISR}_{m,n} \geq \frac{K}{V} 
\frac{\kappa_n}{\kappa_m \kappa_n - 1}
. 
\end{align}
This expression clearly shows how for second-order uncorrelated elliptical sources, the `degree' of non-Gaussianity as expressed by $\kappa$, directly determines the source separation performance.  
In fact, as shown in the following theorem, the same statement holds for second-order correlated elliptical sources.
\begin{theorem}
\label{thm:ISRvsKappa}
If three \glspl{scv} follow distributions from the elliptical family with covariance matrices, $\mx{R}_m = \mx{R}_{m'}$, and $\mx{R}_n$, and $\kappa_{m} \geq \kappa_{m'}$ then $\text{ISR}_{m,n} \leq \text{ISR}_{m',n}$.
\end{theorem}
\begin{IEEEproof}
For elliptically distributed sources, via Lemma \ref{lem:GammaVsInvCov}, we have that $\kappa_m \geq \kappa_{m'} \geq 1$ and $\kappa_n \geq 1$, thus 
\begin{align*}
\kappa_m  \mx{R}_m^{-1} \hadamard \mx{R}_{n} - \kappa_n^{-1} \left( \mx{R}_n^{-1} \hadamard \mx{R}_{m}\right)^{-1} 
&\succeq
\kappa_{m'} \mx{R}_m^{-1} \hadamard \mx{R}_{n} - \kappa_n^{-1} \left( \mx{R}_n^{-1} \hadamard \mx{R}_{m}\right)^{-1}
\\
\left(\kappa_m \mx{R}_m^{-1} \hadamard \mx{R}_{n} - \kappa_n^{-1} \left( \mx{R}_n^{-1} \hadamard \mx{R}_{m}\right)^{-1}\right)^{-1} &\preceq 
\left(\kappa_{m'} \mx{R}_m^{-1} \hadamard \mx{R}_{n} - \kappa_n^{-1} \left( \mx{R}_n^{-1} \hadamard \mx{R}_{m}\right)^{-1}
\right)^{-1},
\end{align*}
and since $\mx{A} \preceq \mx{B}$ implies $\vect{x}^\Tpose \mx{A} \vect{x} \leq \vect{x}^\Tpose \mx{B} \vect{x}, \forall \vect{x}$, and thus $\trace{\mx{A}} \leq \trace{\mx{
B}}$.
\end{IEEEproof}
 
\subsection{\glsentrytext{crlb} for \glsentrytext{ica}}
Another special case, which is of particular interest, is when there is only one dataset, i.e., $K=1$. 
For this case, the expressions above further simplify to the more extensively studied \gls{ica} performance bounds 
\cite{
Tichavsky_CRLBforFastICA_2006,
Ollila_CRLBforICA_2008,
Yeredor_BSSgauss_2010,
Comon_HandbookBSS_2010}. 
If $K=1$, we can replace the \gls{scm} notation with source component notation, i.e., let $\vect{s}_n \in \real^{V}$ be the random vector and the multivariate score function be denoted by $\vectgk{\phi}_m \in \real^{V}$. Then, for this section we have $\mx{R}_n = \expect{\vect{s}_n \vect{s}_n^\Tpose} \in \real^{V \times V}$ 
and 
$\mxgk{\Gamma}_m = \expect{\vectgk{\phi}_m \vectgk{\phi}_m^\Tpose} \in \real^{V \times V}$, from which we observe that for $m \neq n$, $\mxgk{\mathcal{K}}_{m,n} = \mathcal{K}_{m,n} = V^{-1} \trace{\mxgk{\Gamma}_m \mx{R}_n} = V^{-1} \variance{\vectgk{\phi}_m^\Tpose   \vect{s}_n}$.  Also,
\begin{align}
\mx{F}_{m,n} = V\left[\begin{array}{cc}
\frac{1}{V} \variance{\vectgk{\phi}_m^\Tpose   \vect{s}_n} & 1 \\
1     & \frac{1}{V}\variance{\vectgk{\phi}_n^\Tpose   \vect{s}_m}
\end{array}\right] 
.
\end{align}

Two particular subcases in \gls{ica} are of interest.  
The first case is when the samples are \gls{iid} with unit variance, for which
$\mx{R}_n = \mx{I}_V$,
$\mxgk{\Gamma}_m = \expect{\phi_m^2} \mx{I}_V$, and
$\mathcal{K}_{m,n} = \kappa_m$, where $\kappa_m \triangleq \expect{\phi_m^2} \geq 1$.  These simplifications give the same results as in \cite[Eq. 38]{Tichavsky_CRLBforFastICA_2006} and \cite[Thm. 2]{Ollila_CRLBforICA_2008}, namely:
\begin{align}
\text{ISR}_{m,n} \geq \frac{1}{V} 
\left(\kappa_m - \kappa_n^{-1}\right)^{-1} = \frac{1}{V} \frac{\kappa_n}{\kappa_m \kappa_n - 1}
. 
\end{align}

The second subcase of \gls{ica} is for sources with Gaussian sample-to-sample dependence, i.e., $\vect{s_n} \sim \pdfnorm{\vect{0}}{\mx{R}_n \in \real^{V \times V}}$.  Then we have that 
$\mxgk{\Gamma}_m = \mx{R}_m^{-1}$ and $\mathcal{K}_{m,n} = V^{-1} \trace{\mx{R}_m^{-1} \mx{R}_n}$, which corresponds to \cite[Eq. 19]{Yeredor_BSSgauss_2010}.
\section{Examples of Algorithm Performance and CRLB}
In this section, we compare the performance of several \gls{iva} algorithms versus the \gls{icrlb} given in Section \ref{sec:CRLB}.
\subsection{MPE IVA}
\label{sec:MPEIVA}
For our first set of experiments, we consider sources following the \gls{mpe} distribution, an elliptical distribution with $h_e\left(u\right) =  \naturale{-\tfrac{1}{2} u^\beta}$ and normalization constant
$c_K =  \pi^{-K/2} 2^{-K/\left(2\beta\right)} \beta  \Gamma\left(K/2\right)$, where $\beta>0$ is termed the shape parameter.
This distribution possesses a score function which includes the score functions used in both \cite{Kim_ThesisIVA_2006} and \cite{Anderson_IVAMGconf_2010} as special cases.
In this section, we consider \gls{iva-mpe}, where the algorithm was presented in \cite{Anderson_Kotz_2013},  using simulated datasets with \gls{iid} samples from the \gls{mpe} family. 
The performance of \gls{iva-mpe} is compared with the \gls{icrlb} derived in Section \ref{sec:CRLB}.

For this experiment, there are $N=3$ \gls{mpe} \glspl{scv} of dimension $K=5$.  
All the sources use the same shape parameter, $\beta$.  
The covariance matrix associated with each source is randomly picked for the experiment, yet fixed for all trials in the experiment.   
The $k$th entry of each SCV is used as a latent source for the $k$th dataset.
 Entries of the random mixing matrices, $\mx{A}^{[k]}$, are from the standard normal distribution and are randomly selected for each trial.   

We compute the theoretical \gls{icrlb} for \gls{isr} and compare this value with the \gls{isr} achieved using \gls{iva-mpe} with the correct shape parameter for each source.
We then compute the total theoretical normalized \gls{isr}, defined as,
\begin{align*}
\text{ISR} \triangleq \sum_{m=1, n=1, m\neq n}^{N}{V \: \text{ISR}_{m,n}}.
\end{align*}
We compare this theoretical \gls{isr} with the average \gls{isr} computed from $1000$ independent trials of the  algorithm as we vary the number of samples per dataset, $V$.  
  
\begin{figure}
\centering
\includegraphics[trim=8mm 3mm 0 3mm,clip, width=3.7in]{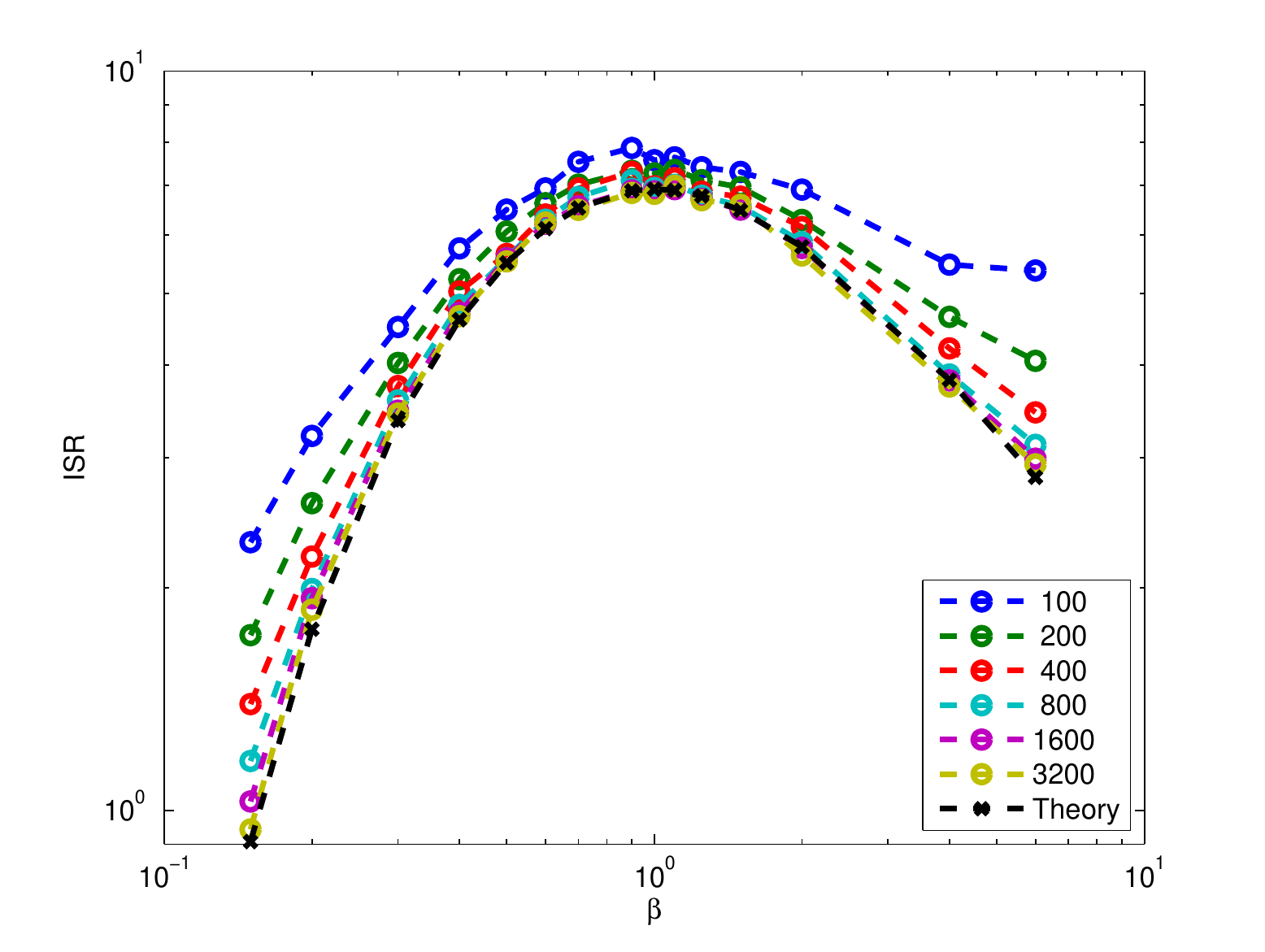} 
\caption{The average \glsentrytext{isr} (of the successful trials) of \glsentrytext{iva-mpe} algorithm for various numbers of \glsentrytext{iid} samples versus the shape parameter of the simulated \glsentrytext{scv} in the \glsentrytext{iid} \glsentrytext{iva} experiment.  The algorithm uses exact knowledge of the shape parameter.  All results are compared with the \glsentrytext{icrlb}.}
\label{fig:fig_MpeVsShape_Corr_ISR}
\end{figure}

Due to the presence of local minima in the \gls{iva} objective function for non-Gaussian sources \cite{Davies_AudioSourceSep_2002},  the algorithm may converge to local minima.  
At local minima, the sources are separated within a dataset but the SCVs are not successfully identified, i.e., the permutation ambiguity is unresolved.
We first compare the \gls{icrlb} for the \gls{isr} with the mean of the \gls{isr} achieved over \emph{successful} trials.  
A trial is deemed successful if the location of the maximum absolute entry in each row of $\mx{G}^{[k]} = \mx{W}^{[k]} \mx{A}^{[k]}$ is unique within each dataset and colocated across the datasets (the former indicates sources are separated within each dataset and the latter indicates if the permutation ambiguity is resolved).
The fraction of trials which are successful increases as $\beta$ decreases and/or as the sample size per dataset increases.
The lowest success rate was 98\%, when $V=100$ and $\beta=6$.
For all other settings the success rate was greater than 99.5\%.
From  \figref{fig:fig_MpeVsShape_Corr_ISR}, the performance of the \gls{iva} algorithm approaches the \gls{icrlb} as the sample size per dataset increases.  

\begin{figure}
\centering
\includegraphics[trim=8mm 3mm 0 3mm,clip, width=3.7in]{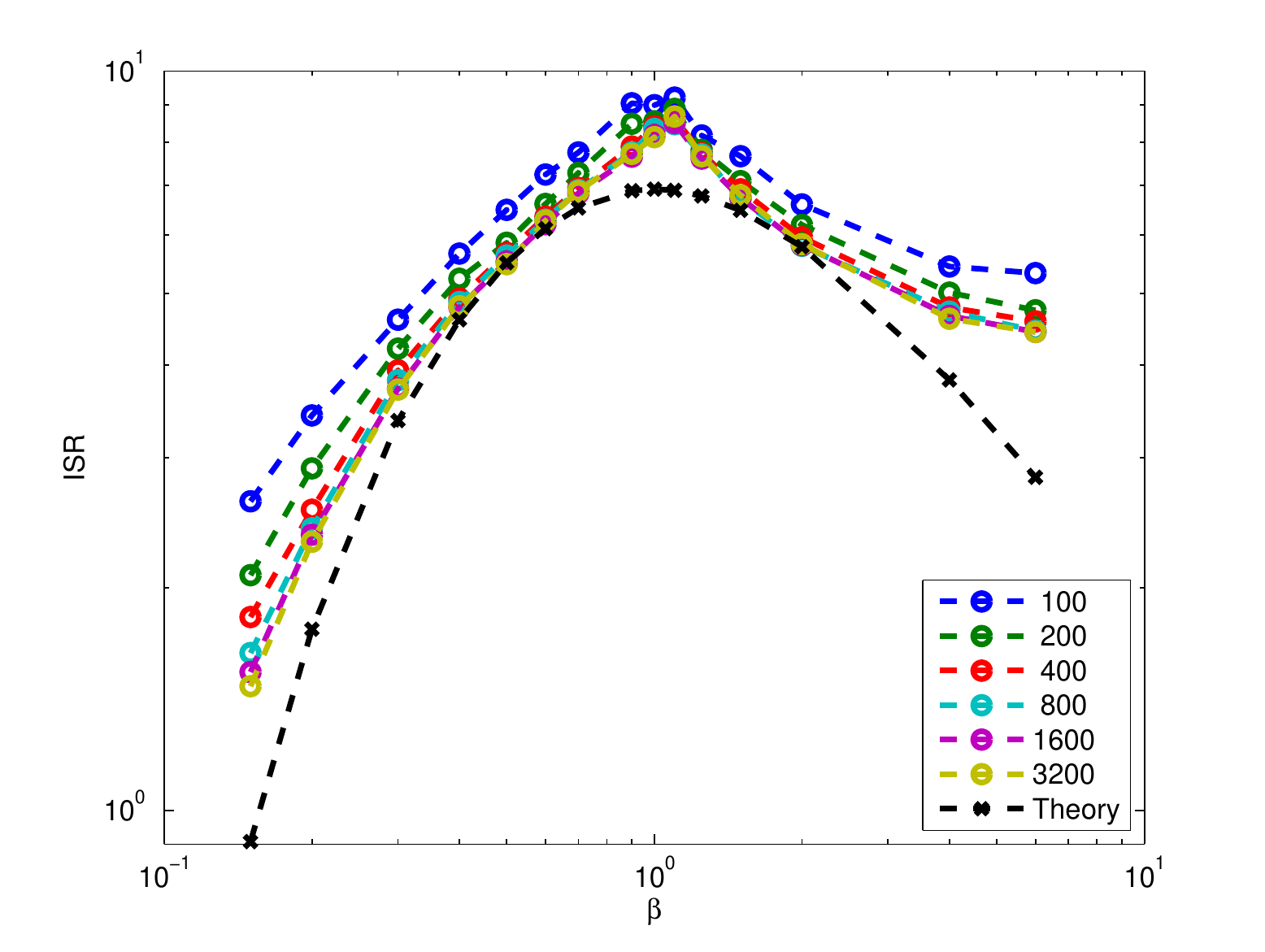}
\caption{The average \glsentrytext{isr} (of the successful trials) of \glsentrytext{iva-mpe} algorithm for various numbers of \glsentrytext{iid} samples versus the shape parameter of the simulated \glsentrytext{scv} in the \glsentrytext{iid} \glsentrytext{iva} experiment.  The algorithm selects from one of two shape parameters, $\beta \in \left\{0.5,2.0\right\}$, and thus does not use exact knowledge of the shape parameter.  All results are compared with the \glsentrytext{icrlb}.
}
\label{fig:fig_MpeVsShape_Corr_2shapes_ISR}
\end{figure}

We also show in \figref{fig:fig_MpeVsShape_Corr_2shapes_ISR}---for the same experiment described above---the performance of the \gls{iva-mpe} when the algorithm selects between one of two shape parameters ($\beta \in \left\{0.5, 2.0\right\}$) according to which shape parameter provides the lowest cost.

\begin{figure}
\centering
\includegraphics[trim=0mm 0mm 0 0mm,clip, width=3.7in]{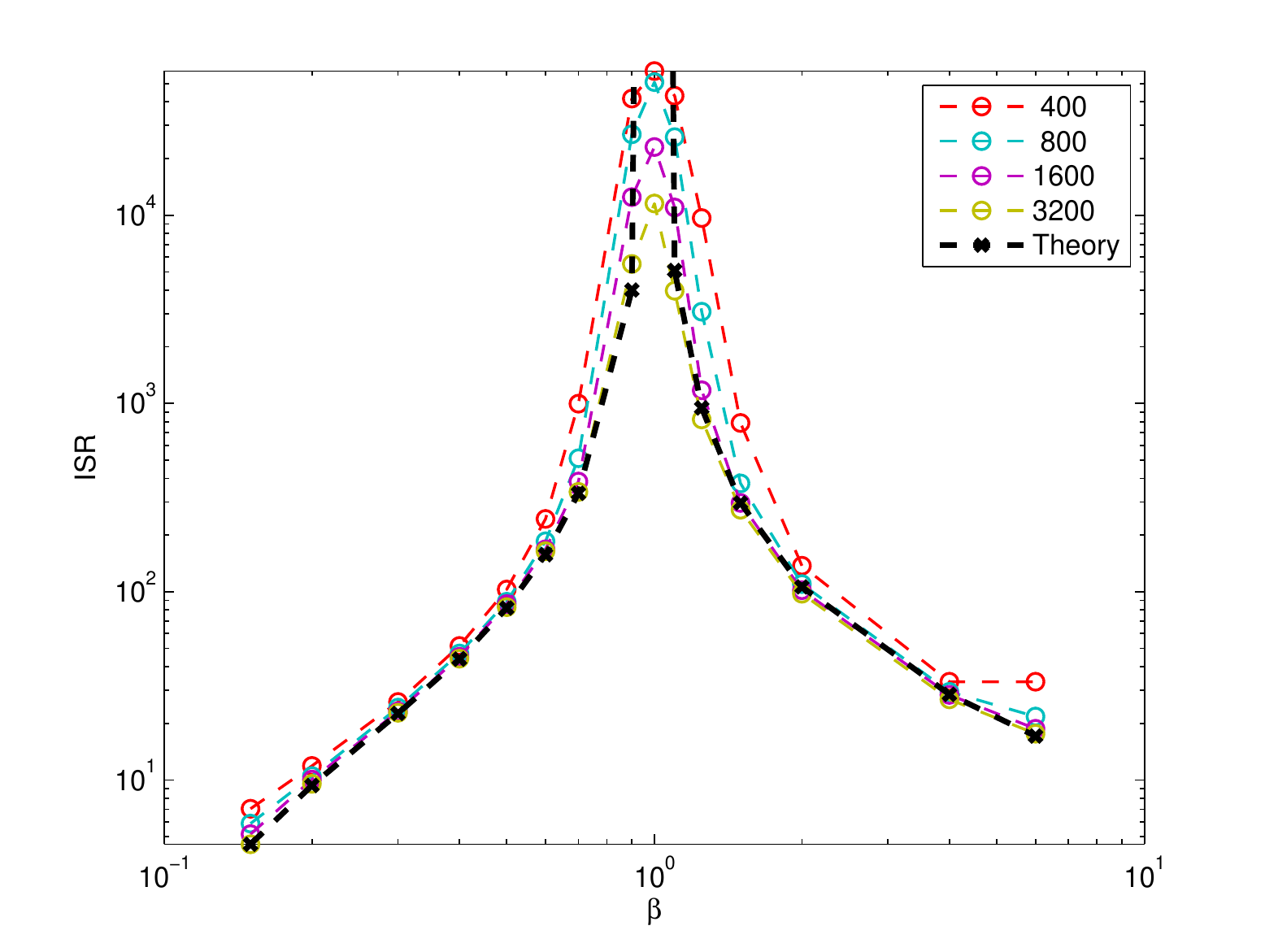}
\caption{The \glsentrytext{icrlb} theory for \glsentrytext{isr} as the shape parameter, $\beta$, varies is compared with the median \glsentrytext{isr} of all $1000$ trials for different numbers of \glsentrytext{iid} samples, $V$.}
\label{fig:MPE_crlb_Vs_shape_uncorr_VaryT}
\end{figure}

In another experiment, we use the same parameters as before except now the \glspl{scv} each have identity covariance matrices.
For this experiment, there are nonidentifiable conditions as $\beta \rightarrow 1$, thus we compare the \gls{icrlb} for the \gls{isr} with the median rather than the mean.
From \figref{fig:MPE_crlb_Vs_shape_uncorr_VaryT}, the performance of the \gls{iva} algorithm  approaches the \gls{icrlb} as the sample size per dataset increases.

In both \figref{fig:fig_MpeVsShape_Corr_ISR} and \figref{fig:MPE_crlb_Vs_shape_uncorr_VaryT}, the \gls{icrlb} follows the behavior predicted by Theorems \ref{thm:ivaiid_id}, \ref{thm:ISRbound}, and \ref{thm:ISRvsKappa}.  Namely, the \gls{icrlb} is infinite when sources are Gaussian and $\mx{R}_n = \mx{I}$ for all sources; the maximum \gls{isr} occurs when sources are Gaussian ($\beta = 1$); and as $\beta$ moves `away' from one the non-Gaussianity measure $\kappa$ increases, which yields better source separation, i.e., lower \gls{isr}.

\subsection{Orthogonal Generalized Joint Diagonalization with Second-Order Lags}
In this section, we consider the effect of sample dependency.  To the best of our knowledge, there is only one algorithm in the \gls{iva} framework that accounts for sample-to-sample dependence, namely \gls{jdiag-sos} as given in \cite{XiLinLi_JointDiagOrthoProcustes_2011}. 
The performance of \gls{jdiag-sos} is compared with the \gls{icrlb} derived in Section \ref{sec:CRLB}.

All the sources are a vector moving average of \gls{iid}  Gaussian samples, i.e., 
\begin{align}
\vect{s}_n \left(v\right) = \sum_{l=0}^{L-1} \mx{B}_l \vect{z}\left(v-l\right),
\end{align}
where $\vect{z} \sim \pdfnorm{\vect{0}}{\mx{I}_K}$ and $\mxelement{\mx{B}_l}{k_1}{k_2} \sim \pdfnorm{0}{1}$. 
For this experiment, there are $N=3$ sources for $K=3$ datasets, each with $V=1000$ samples and $L=4$. 
Entries of the random mixing matrices, $\mx{A}^{[k]}$, are from the standard normal distribution and are randomly selected for each trial.   

We compute the theoretical \gls{icrlb} for \gls{isr} assuming the data was generated with $L=1, \ldots, 4$.  
Since $L=4$ for the data, the performance bound is shown to decrease until the lag is 3.
The performance bound for $L=4$ is shown for lags greater than 3. 
We compare the performance bounds with the average over $100$ independent trials of the \gls{isr}  achieved using \gls{jdiag-sos} with various lags.
Due to \gls{jdiag-sos} estimating orthogonal demixing matrices there exists a noticeable difference between the \gls{icrlb} for \gls{isr} and the observed \gls{isr}.

\begin{figure}
\centering
\includegraphics[trim=6mm 3mm 0 4mm,clip, height=2.7in]{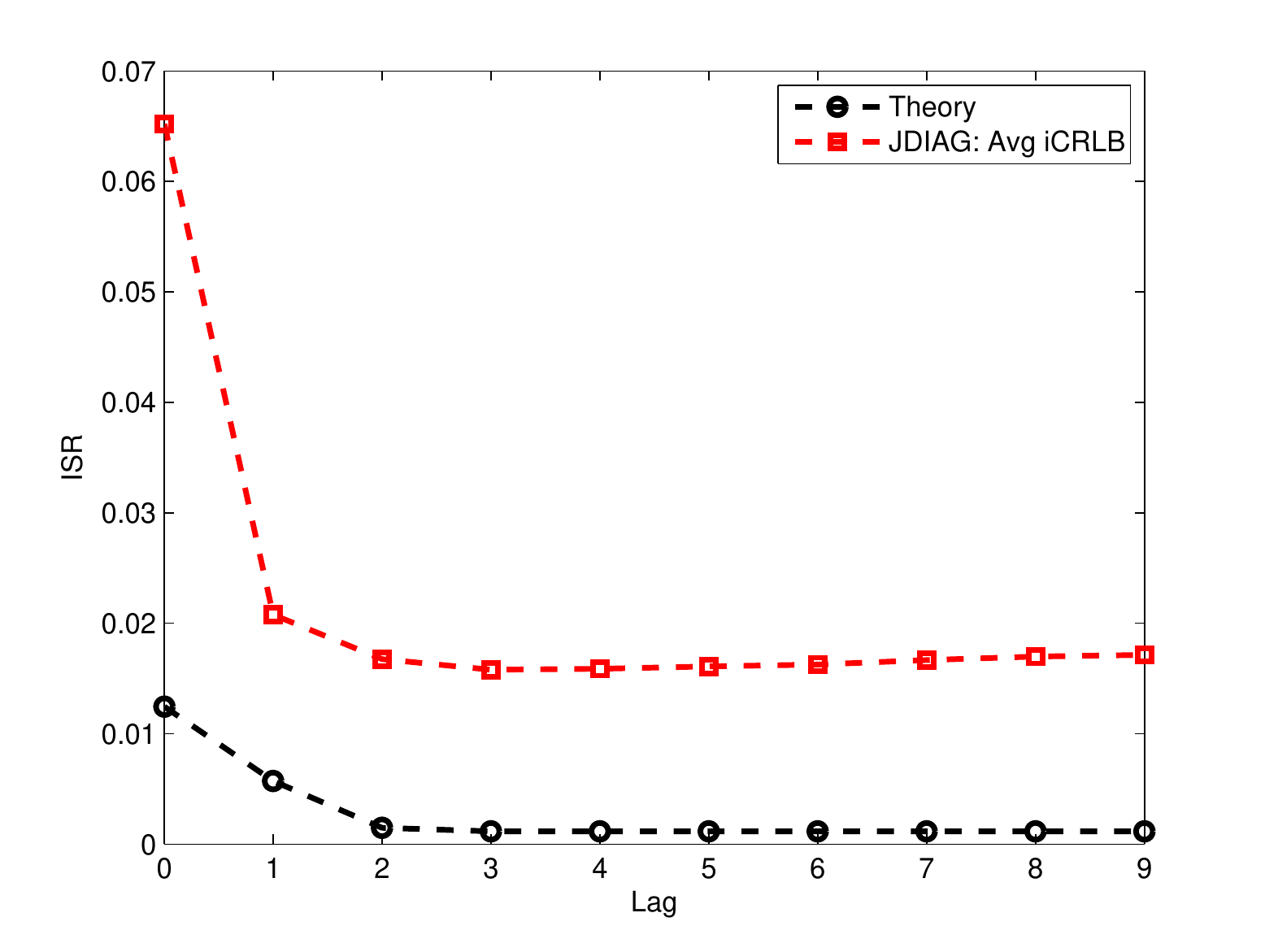}
\caption{The average \glsentrytext{isr} for 100 trials  by \glsentrytext{jdiag-sos}(L).  The number of lags used by \glsentrytext{jdiag-sos} is varied from 0 to 9 ($L=1, \ldots, 10$).  The \glsentrytext{icrlb} is shown assuming at most lag = 3.}
\label{fig:fig_IVAwithLags}
\end{figure}

\section{Conclusion}
The use of \gls{iva} for the separation of multiple datasets concurrently has been a more recent development within the general \gls{bss} literature.
A variety of algorithms have been developed that are essentially the multivariate extensions of \gls{ica} algorithms which take into account the dependence of sources between datasets in a variety of ways.
There are three principal reasons for using these algorithms (versus just using \gls{ica} individually on each dataset).
First, to increase the set of sources which can be identified.  
Second, to automatically `align' dependent sources.
Third, to maximize the achievable source separation.
In this work, we have given the larger set of sources which can be identified by \gls{iva}, proven when the estimated sources can be `aligned', and provided the bound on achievable source separation using \gls{iva}.
These results are achieved for an \gls{iva} that accounts for linear and nonlinear dependence of sources across datasets, non-Gaussianity, and sample-to-sample dependence. 
It is clear that \gls{iva} bridges the gap between \gls{cca} and \gls{ica}.

It will be interesting for future work to consider the additional diversity of complex-valued sources which are improper or noncircular.
Additionally, our work will be useful for assessing the performance of future algorithms which account for sample dependency in an \gls{iva} framework.

\appendices
\section{Derivation of \glsentrytext{iva} \glsentrytext{fim}}\label{sec:DeriveFIM}
Here we derive the \gls{fim} of \eqref{eq:JBSS_Likelihood} \gls{wrt} $\mxgk{\mathcal{W}}$.  The $KN^2$ parameters result in $KN^2 \times KN^2$ dimension \gls{fim} with the entry associated with $w^{[k_1]}_{m_1,n_1}$ and $w^{[k_2]}_{m_2,n_2}$ given by \eqref{eq:FIM_W}.

For the computations to follow it is useful to observe that,
\begin{align}
\frac{\partial \log\abs{\det \mx{W}^{\left[l\right]}}}{\partial w^{[k]}_{m,n}}
& =
\delta_{l,k} \vect{e}_m^\Tpose \left(\mx{W}^{[k]}\right)^{-\Tpose} \vect{e}_n
\end{align}

\begin{align*}
\mx{Y}_{m}^\Tpose = \left[\left(\mx{X}^{[1]}\right)^\Tpose \vect{w}_m^{[1]}, \ldots, \left(\mx{X}^{[K]}\right)^\Tpose \vect{w}_m^{[K]} \right] \in \real^{K \times V},
\end{align*}

\begin{align}
\frac{\partial \mx{Y}_{l}}{\partial w^{[k]}_{m,n}}
& =	
\delta_{l,m}\Diag{\vect{e}_k}\mx{X}_n \in \real^{K \times V},
\end{align}
and
\begin{align}
\frac{\partial \log \left(p_{m}\left(\mx{Y}_m\right)\right)}{\partial w^{[k]}_{m,n}} 
 & 
=
\trace{
\frac{\partial \log \left(p_{m}\left(\mx{Y}_m\right)\right) }{ \partial \mx{Y}_m^\Tpose} 
\frac{\partial\mx{Y}_m}{\partial w^{[k]}_{m,n}}
}
\label{eq:partialLogP_dW_step1}
\\ & 
=
-\trace{\mxgk{\Phi}_m^\Tpose \Diag{\vect{e}_k}\mx{X}_n}
\\ & 
=
-\left(
\vectgk{\phi}^{[k]}_m
\right)^\Tpose \vect{x}^{[k]}_n 
.
\end{align}
Note that \eqref{eq:partialLogP_dW_step1} is due to applying the chain rule given in \cite[Sect. 2.8.1]{Petersen_MatrixCookboox_2008.pdf}.
Thus the gradient of the likelihood function in \eqref{eq:JBSS_Likelihood} is
\begin{align*}
\frac{\partial \mathcal{L}\left(\mxgk{\mathcal{W}}\right)}{\partial w^{[k]}_{m,n}}
=
-\left(\vectgk{\phi}_m^{[k]}\right)^\Tpose \vect{x}^{[k]}_n
+ V w_{n,m}^{-[k]},
\end{align*}
where $w^{-[k]}_{m,n}$ is the entry in the $m$th row and $n$th column $\left(\mx{W}^{[k]}\right)^{-1}$.

\begin{figure}
\centering
\includegraphics[width=3.5in]{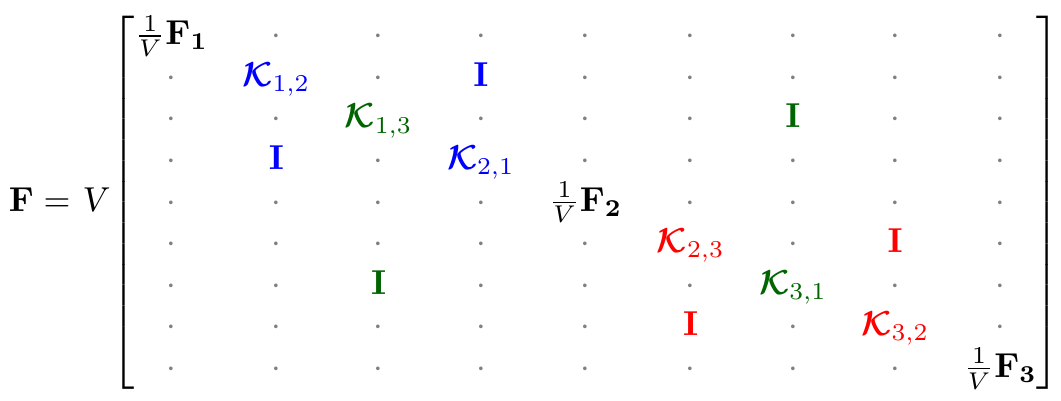}
\caption{Form of \glsentrytext{fim} when $N=3$ sources.  All entries are of $K \times K$ matrices and we use $\cdot$ denote zero blocks.  The entries of \glsentrytext{fim} associated with $\mx{F}_{1,2}$, $\mx{F}_{1,3}$, and $\mx{F}_{2,3}$ are indicated by blue, green, and red, respectively.}
\label{fig:fig_FIM_N3}
\end{figure}

Letting $\mx{A} = \mx{W} = \mx{I}$ we have the \gls{fim} of interest with entries given by
\begin{align}
\begin{split}
\left[\mx{F}\right]^{k_1,m_1,n_1}_{k_2,m_2,n_2} 
 \triangleq &
\left. \left[\mx{F}\left(\mxgk{\mathcal{W}}\right)\right]^{k_1,m_1,n_1}_{k_2,m_2,n_2} 
\right|_{\mx{A} = \mx{I}, \mx{W}=\mx{I}}
\\
 = &
\expect{\left((\vectgk{\phi}_{m_1}^{[k_1]}\right)^\Tpose \vect{s}^{[k_1]}_{n_1} \left(\vect{s}^{[k_2]}_{n_2}\right)^\Tpose
\vectgk{\phi}_{m_2}^{[k_2]}}
+ V^2 \delta_{m_1,n_1} \delta_{m_2,n_2}
\\
&
- V \expect{\left(\vectgk{\phi}_{m_2}^{[k_2]}\right)^\Tpose \vect{s}^{[k_2]}_{n_2}} \delta_{m_1,n_1}
\\
&
- V \expect{\left(\vectgk{\phi}_{m_1}^{[k_1]}\right)^\Tpose \vect{s}^{[k_1]}_{n_1}} \delta_{m_2,n_2}
\\
= & 
\expect{\left(\vectgk{\phi}_{m_1}^{[k_1]}\right)^\Tpose \vect{s}^{[k_1]}_{n_1} \left(\vect{s}^{[k_2]}_{n_2}\right)^\Tpose
\vectgk{\phi}_{m_2}^{[k_2]}}
- V^2 \delta_{m_1,n_1} \delta_{m_2,n_2},
\end{split}
\end{align}
where the following expression holds,
$\expect{\vect{s}^{[k_1]}_{n} \left(\vectgk{\phi}_{m}^{[k_2]}\right)^\Tpose} = \delta_{k_1,k_2} \delta_{m,n} \mx{I}_V$, see \cite{Comon_HandbookBSS_2010}.  
Since, by assumption, both $\expect{\vectgk{\phi}_m^{[k]}} = \vect{0}$ and $\expect{\vect{s}_m^{[k]}} = \vect{0}$, then it is 
true that $\left[\mx{F}\right]^{k_1,m_1,n_1}_{k_2,m_2,n_2}=0$ when one of the entries in $\left(m_1,n_1,m_2,n_2\right)$ is unique.  It is also zero when $m_1=n_1 \neq m_2 = n_2$, i.e., $\expect{\left(\vectgk{\phi}_{m_1}^{[k_1]}\right)^\Tpose \vect{s}^{[k_1]}_{m_1}}
\expect{\left(\vect{s}^{[k_2]}_{m_2}\right)^\Tpose
\vectgk{\phi}_{m_2}^{[k_2]}}
- V^2 = V^2-V^2 = 0$.
Thus, there are only three nonzero cases to consider:
\begin{align*}
\left[\mx{F}\right]^{k_1,m_1,n_1}_{k_2,m_2,n_2}
& =
\begin{cases}
V \left(\mathcal{K}_{m_1,m_1}^{[k_1,k_2]} 
- V\right)
& m_1 = n_2 = m_2 = n_1
\\
V \mathcal{K}_{m_1,n_1}^{[k_1,k_2]} 
 & m_1=m_2 \neq n_1 = n_2
\\
V \delta_{k_1,k_2} 
& m_1 = n_2 \neq m_2 = n_1
\\
0 & \textrm{otherwise},
\end{cases}
\end{align*}
where 
$\mathcal{K}_{m,n}^{[k_1,k_2]} \triangleq \frac{1}{V} \expect{\left(\vectgk{\phi}_{m}^{[k_1]}\right)^\Tpose \vect{s}^{[k_1]}_{n} \left(\vect{s}^{[k_2]}_{n}\right)^\Tpose
\vectgk{\phi}_{m}^{[k_2]}} = \frac{1}{V} \trace{\expect{
\vectgk{\phi}_{m}^{[k_2]} \left(\vectgk{\phi}_{m}^{[k_1]}\right)^\Tpose
\vect{s}_{n}^{[k_1]} \left(\vect{s}^{[k_2]}_{n}\right)^\Tpose
}}$ is the $\left(k_1,k_2\right)$ entry of $\mxgk{\mathcal{K}}_{m,n}$.  
The form of this matrix (e.g., see \figref{fig:fig_FIM_N3}) is the block-matrix extension of that for the single dataset \gls{fim} given in  \cite{Loesch_ComplexCrlbICA_2013}.

There exists a permuted \gls{fim} in which there are $N + N\left(N-1\right)/2$ nonzero matrices along the diagonal, i.e.,
\begin{align}
 \mx{F}  &= \left[\begin{array}{cc}
\oplus_{n=1}^N \mx{F}_n & \mx{0} \\
\mx{0} & \oplus_{m=1,n=m+1}^{N,N} \mx{F}_{m,n}
\end{array}
\right].
\label{eq:app:IVAFIM}
\end{align}
The submatrices are given by
\begin{align}
\mx{F}_{n} & \triangleq 
\variance{\diag{\mxgk{\Phi}_n \mx{S}_n^\Tpose - \mx{I}_V}} 
= V \left(\mxgk{\mathcal{K}}_{n,n} -V \mx{1}_{K \times K}\right)
\label{eq:app:IVAFIM_n}
\end{align}
and
\begin{align}
\mx{F}_{m,n} & \triangleq 
\covariance{\left[\begin{array}{cc}
\diag{\mxgk{\Phi}_m \mx{S}_n^\Tpose}  \\
\diag{\mxgk{\Phi}_n \mx{S}_m^\Tpose}
\end{array}\right]}
=V \left[\begin{array}{cc}
\mxgk{\mathcal{K}}_{m,n} & \mx{I}_K \\
\mx{I}_K       & \mxgk{\mathcal{K}}_{n,m}
\end{array}\right] 
\label{eq:app:IVAFIM_mn},
\end{align}
where $\mx{F}_n \in \real^{K \times K}$ and $\mx{F}_{m,n} \in  \real^{2K \times 2K}$.
It is also useful to note that for $1 \leq m \neq n \leq N$ we have
$
\mathcal{K}_{m,n}^{[k_1,k_2]} 
=
\frac{1}{V}
\trace{\mxgk{\Gamma}_{m}^{[k_2,k_1]} \mx{R}_{n}^{[k_1,k_2]}}
$,
where
$\mx{R}_{n}^{[k_1,k_2]} \triangleq \expect{\vect{s}_n^{[k_1]} \left(\vect{s}_n^{[k_2]}\right)^\Tpose} \in \real^{V \times V}$ and
$\mxgk{\Gamma}_{n}^{[k_1,k_2]} 
\triangleq 
\expect{\vectgk{\phi}^{[k_1]}_n \left(\vectgk{\phi}^{[k_2]}_n\right)^\Tpose}
 \in \real^{V \times V}$. 

\section{Score Function Covariance Matrix for Elliptical Distributions}
\label{sec:EllipticalGammaMx}
In this appendix, we show that the score function covariance matrix, $\mxgk{\Gamma} = \expect{\vectgk{\phi} \vectgk{\phi}^\Tpose}$, is a scalar multiple of the inverse of the covariance matrix for all elliptical distributions defined by \eqref{eq:ellipticalpdf}. 
We begin by letting $\vect{z} = \mxgk{\Sigma}^{-1/2} \vect{x}$ so that $
p_{\vect{z}}\left(\vect{z} \right)
=
\abs{\det{\mxgk{\Sigma}^{-1/2}}}^{-1}p_{\vect{x}}\left(\mxgk{\Sigma}^{1/2} \vect{z}\right)
=
c_K h_e\left(\vect{z}^\Tpose  \vect{z}\right)
$, which results in $
\mxgk{\Gamma}
 = 
  \mxgk{\Sigma}^{-1/2} \expect{g^2\left(\vect{z}^\Tpose  \vect{z} \right) \vect{z} \vect{z}^\Tpose } \mxgk{\Sigma}^{-1/2}
$. 
To compute the expectation requires the following multivariate integral to be evaluated:
\begin{align}
\expect{g^2\left(\vect{z}^\Tpose \vect{z} \right) z_l z_k}
&=
\int_{-\infty}^\infty 
g^2\left(\vect{z}^\Tpose \vect{z} \right) z_l z_k p\left(\vect{z}\right) d \vect{z}.
\label{eq:ExpectKappa}
\end{align}
We use a transformation of variables utilized for similar problems in \cite{Muirhead_MultivariateStatisticalTheory_2005, Aulogiaris_Kotz_2004}, namely,
\begin{align}
z_1 &= r \prod_{k=1}^{K-1} \sin \theta_k 
\\
z_j &= r \left(\prod_{k=1}^{K-j} \sin \theta_k \right) \cos \theta_{K-j+1}, 2\leq j \leq K-1
\\
z_K &= r \cos \theta_1
\end{align}
where $0 < \theta_j \leq \pi$, $j=1,\ldots,K-2$, $0 < \theta_{K-1} \leq 2\pi$, $0 < r \leq \infty$.  
By noting that $\vect{z}^\Tpose \vect{z} = r^2$ and the Jacobian of the transformation from $\vect{z}$ to $\left[\theta_1 \ldots \theta_{K-1} \: r\right]^\Tpose$ is $r^{K-1} \sin^{K-2}\theta_1
\sin^{K-3}\theta_2 \cdots \sin\theta_{K-2} 
= r^{K-1} \prod_{k=1}^{K-2} \left(\sin \theta_k\right)^{K-1-k}$, we have $p\left(r\right)
=
c_K h_e\left(r^2\right)
$.
  
There are two cases, $l=k$  and $l\neq k$, required to evaluate \eqref{eq:ExpectKappa}.  Let us consider the former first,
\begin{align}
\expect{g^2\left(\vect{z}^\Tpose \vect{z} \right) z_1^2}
&=
\expect{ g^2(r^2) r^{K+1}} 
\frac{2\pi^{K/2}}{K
\Gamma\left(K/2\right)}
,
\end{align}
where we have made use of $\int_0^{\pi } \sin^n \theta d\theta = \sqrt{\pi } \Gamma\left[\left(n+1\right)/2\right] / \Gamma\left[\left(n+2\right)/2\right]$ when $n \geq 1$.

Now for the off-diagonal terms, e.g., when $K=2$,
$
\expect{g^2\left(\vect{z}^\Tpose \vect{z} \right) z_1 z_2} = \int_{0}^\infty \int_{0}^{2\pi}  g^2(r^2) p(r) r \cos \theta \sin \theta d\theta dr = 0
$, where  we have used the following $\int_0^{n\pi} \cos \left(\theta\right) \sin^n \left(\theta\right) d\theta = 0$ when $n \in \mathbb{N}^*$.
The result holds for the more general case when $K>2$ and $l \neq k$ and we arrive at the final expression of 
$
 \expect{\vectgk{\phi} \vectgk{\phi}^\Tpose} 
 = 
\mxgk{\Sigma}^{-1/2} \expect{g^2\left(\vect{z}^\Tpose  \vect{z} \right) \vect{z} \vect{z}^\Tpose } \mxgk{\Sigma}^{-1/2}
= 
\expect{g^2\left(\vect{z}^\Tpose  \vect{z} \right) \vect{z} \vect{z}^\Tpose } \mxgk{\Sigma}^{-1}
 =
 \kappa \mx{R}^{-1}, K \geq 2
 $
 , 
where $\kappa \triangleq  \expect{ g^2(r^2) r^{K+1}} 
\tfrac{2\pi^{K/2}}{K
\Gamma\left(K/2\right)} \rho$.



\begin{thebibliography}{10}
\providecommand{\url}[1]{#1}
\csname url@samestyle\endcsname
\providecommand{\newblock}{\relax}
\providecommand{\bibinfo}[2]{#2}
\providecommand{\BIBentrySTDinterwordspacing}{\spaceskip=0pt\relax}
\providecommand{\BIBentryALTinterwordstretchfactor}{4}
\providecommand{\BIBentryALTinterwordspacing}{\spaceskip=\fontdimen2\font plus
\BIBentryALTinterwordstretchfactor\fontdimen3\font minus
  \fontdimen4\font\relax}
\providecommand{\BIBforeignlanguage}[2]{{%
\expandafter\ifx\csname l@#1\endcsname\relax
\typeout{** WARNING: IEEEtran.bst: No hyphenation pattern has been}%
\typeout{** loaded for the language `#1'. Using the pattern for}%
\typeout{** the default language instead.}%
\else
\language=\csname l@#1\endcsname
\fi
#2}}
\providecommand{\BIBdecl}{\relax}
\BIBdecl

\bibitem{Ica_Book_2001}
A.~Hyv{\"a}rinen, J.~Karhunen, and E.~Oja, \emph{Independent Component
  Analysis}.\hskip 1em plus 0.5em minus 0.4em\relax Wiley-Interscience, 2001.

\bibitem{Comon_HandbookBSS_2010}
P.~Comon and C.~Jutten, \emph{Handbook of Blind Source Separation: Independent
  Component Analysis and Applications}, 1st~ed.\hskip 1em plus 0.5em minus
  0.4em\relax Academic Press, 2010.

\bibitem{Lee_IVAfMRI_2008}
J.-H. Lee, T.-W. Lee, F.~A. Jolesz, and S.-S. Yoo, ``Independent vector
  analysis ({IVA}): Multivariate approach for {fMRI} group study,''
  \emph{NeuroImage}, vol.~40, no.~1, pp. 86--109, 2008.

\bibitem{Li_MCCA_2009}
Y.-O. Li, T.~Adal{\i}, W.~Wang, and V.~D. Calhoun, ``Joint blind source
  separation by multiset canonical correlation analysis,'' \emph{IEEE Trans.
  Signal Process.}, vol.~57, no.~10, pp. 3918--3929, Oct. 2009.

\bibitem{Kim_RealtimeIVA_2010}
T.~Kim, ``Real-time independent vector analysis for convolutive blind source
  separation,'' \emph{Circuits and Systems I: Regular Papers, IEEE Transactions
  on}, vol.~57, no.~7, pp. 1431--1438, Jul. 2010.

\bibitem{Kettenring_MCCA_1971}
J.~R. Kettenring, ``Canonical analysis of several sets of variables,''
  \emph{Biometrika}, vol.~58, no.~3, pp. 433--451, 1971.

\bibitem{Young_NonlinearCCA_1976}
F.~W. Young, J.~{De Leeuw}, and Y.~Takane,
  ``\BIBforeignlanguage{English}{Regression with qualitative and quantitative
  variables: An alternating least squares method with optimal scaling
  features},'' \emph{\BIBforeignlanguage{English}{Psychometrika}}, vol.~41, pp.
  505--529, 1976.

\bibitem{Anderson_IVAG_2012}
M.~Anderson, T.~Adal{\i}, and X.-L. Li, ``Joint blind source separation of
  multivariate {G}aussian sources: Algorithms and performance analysis,''
  \emph{IEEE Trans. Signal Process.}, vol.~60, no.~4, pp. 1672--1683, Apr.
  2012.

\bibitem{Hotelling_CCA_1936}
H.~Hotelling, ``Relations between two sets of variates,'' \emph{Biometrika},
  vol.~28, no. 3/4, pp. 321--377, 1936.

\bibitem{Anderson_IVAMGconf_2010}
M.~Anderson, X.-L. Li, and T.~Adal{\i}, ``Nonorthogonal independent vector
  analysis using multivariate {G}aussian model,'' in \emph{Latent Variable
  Analysis and Signal Separation}, ser. Lecture Notes in Computer
  Science.\hskip 1em plus 0.5em minus 0.4em\relax Springer Berlin / Heidelberg,
  2010, vol. 6365, pp. 354--361.

\bibitem{Via_ML_IVA_MLSP2011}
J.~V\'{i}a, M.~Anderson, X.-L. Li, and T.~Adal{\i}, ``A maximum likelihood
  approach for independent vector analysis of {G}aussian data sets,'' in
  \emph{IEEE International Workshop on Machine Learning for Signal Processing
  (MLSP 2011)}, Beijing, China, Sep. 2011.

\bibitem{XiLinLi_JointDiagOrthoProcustes_2011}
X.-L. Li, T.~Adal{\i}, and M.~Anderson, ``Joint blind source separation by
  generalized joint diagonalization of cumulant matrices,'' \emph{Signal
  Process.}, vol.~91, no.~10, pp. 2314--2322, Oct. 2011.

\bibitem{XiLinLi_JointDiagGradient_2010}
X.-L. Li, M.~Anderson, and T.~Adal{\i}, ``Second and higher-order correlation
  analysis of multiset multidimensional variables by joint diagonalization,''
  in \emph{Latent Variable Analysis and Signal Separation}, ser. Lecture Notes
  in Computer Science.\hskip 1em plus 0.5em minus 0.4em\relax Springer Berlin /
  Heidelberg, 2010, vol. 6365, pp. 197--204.

\bibitem{deLeeuw_GifiSystem_1984}
J.~{d}e Leeuw, ``The {G}ifi-system of nonlinear multivariate analysis,''
  \emph{Data Analysis and Informatics}, vol. III, pp. 415--424, 1984.

\bibitem{Gifi_Book_1990}
A.~Gifi, \emph{Nonlinear multivariate analysis}.\hskip 1em plus 0.5em minus
  0.4em\relax New York: Wiley, 1990.

\bibitem{Yin_ICCA_2004}
X.~Yin, ``Canonical correlation analysis based on information theory,''
  \emph{Journal of Multivariate Analysis}, vol.~91, no.~2, pp. 161--176, 2004.

\bibitem{Akaho_KernelCCA_2001}
S.~Akaho, ``A kernel method for canonical correlation analysis,'' in
  \emph{International Meeting on Psychometric Society (IMPS2001)}, 2001.

\bibitem{Melzer_NonlinearCCA_2001}
T.~Melzer, M.~Reiter, and H.~Bischof, ``\BIBforeignlanguage{English}{Nonlinear
  feature extraction using generalized canonical correlation analysis},'' in
  \emph{\BIBforeignlanguage{English}{Artificial Neural Networks - ICANN 2001}},
  ser. Lecture Notes in Computer Science, G.~Dorffner, H.~Bischof, and
  K.~Hornik, Eds.\hskip 1em plus 0.5em minus 0.4em\relax Springer Berlin
  Heidelberg, 2001, vol. 2130, pp. 353--360.

\bibitem{Todros_MTCCA_2012}
K.~Todros and A.~O. Hero, ``On measure transformed canonical correlation
  analysis,'' \emph{IEEE Trans. Signal Process.}, vol.~60, no.~9, pp.
  4570--4585, Sep. 2012.

\bibitem{Kim_IVA_2006}
T.~Kim, T.~Eltoft, and T.-W. Lee, ``Independent vector analysis: an extension
  of {ICA} to multivariate components,'' in \emph{Independent Component
  Analysis and Blind Signal Separation}, ser. Lecture Notes in Computer
  Science.\hskip 1em plus 0.5em minus 0.4em\relax Springer Berlin / Heidelberg,
  2006, vol. 3889, pp. 165--172.

\bibitem{Kim_IVAasilomar_2006}
T.~Kim, I.~Lee, and T.-W. Lee, ``Independent vector analysis: Definition and
  algorithms,'' in \emph{Proc. of 40th Asilomar Conference on Signals, Systems,
  and Computers}, Oct. 2006, pp. 1393--1396.

\bibitem{Hiroe_AlmostIVA_2006}
A.~Hiroe, ``Solution of permutation problem in frequency domain {ICA}, using
  multivariate probability density functions,'' in \emph{Independent Component
  Analysis and Blind Signal Separation}, ser. Lecture Notes in Computer
  Science, J.~Rosca, D.~Erdogmus, J.~C. Pr{\'{i}}ncipe, and S.~Haykin,
  Eds.\hskip 1em plus 0.5em minus 0.4em\relax Springer Berlin Heidelberg, 2006,
  vol. 3889, pp. 601--608.

\bibitem{Smaragdis_ConvolvedBSS_1998}
P.~Smaragdis, ``Blind separation of convolved mixtures in the frequency
  domain,'' \emph{Neurocomputing}, vol.~22, no. 1--3, pp. 21--34, 1998.

\bibitem{Phlypo_CDA_2013}
R.~Phlypo, ``Jacobi iterations for canonical dependence analysis,''
  \emph{Signal Processing}, vol.~93, no.~1, pp. 185--197, 2013.

\bibitem{Horn_MatrixAnalysis_1985}
R.~A. Horn and C.~R. Johnson, \emph{Matrix Analysis}.\hskip 1em plus 0.5em
  minus 0.4em\relax Cambridge: Cambridge University Press, 1985.

\bibitem{Kim_IVA_2007}
T.~Kim, H.~T. Attias, S.-Y. Lee, and T.-W. Lee, ``Blind source separation
  exploiting higher-order frequency dependencies,'' \emph{IEEE Trans. Audio
  Speech Lang. Process.}, vol.~15, no.~1, pp. 70--79, Jan. 2007.

\bibitem{Cover_InfoTheoryBook_2006}
T.~M. Cover and J.~A. Thomas, \emph{Elements of Information Theory}.\hskip 1em
  plus 0.5em minus 0.4em\relax Wiley-Interscience, 2006.

\bibitem{Cardoso_Equivariant_1996}
J.-F. Cardoso and B.~H. Laheld, ``Equivariant adaptive source separation,''
  \emph{IEEE Trans. Signal Process.}, vol.~44, no.~12, pp. 3017--3030, Dec.
  1996.

\bibitem{Tichavsky_CRLBforFastICA_2006}
P.~Tichavsk{\'{y}}, Z.~Koldovsk{\'{y}}, and E.~Oja, ``Performance analysis of
  the {FastICA} algorithm and {C}ram{\'{e}}r-{R}ao bounds for linear
  independent component analysis,'' \emph{IEEE Trans. Signal Process.},
  vol.~54, no.~4, pp. 1189--1203, Apr. 2006.

\bibitem{Ollila_CRLBforICA_2008}
E.~Ollila, K.~Hyon-Jung, and V.~Koivunen, ``Compact {C}ram{\'{e}}r-{R}ao bound
  expression for independent component analysis,'' \emph{IEEE Trans. Signal
  Process.}, vol.~56, no.~4, pp. 1421--1428, Apr. 2008.

\bibitem{Yeredor_BSSgauss_2010}
A.~Yeredor, ``Blind separation of {G}aussian sources with general covariance
  structures: Bounds and optimal estimation,'' \emph{IEEE Trans. Signal
  Process.}, vol.~58, no.~10, pp. 5057--5068, Oct. 2010.

\bibitem{Loesch_ComplexCrlbICA_2013}
B.~Loesch and B.~Yang, ``Cramer-rao bound for circular and noncircular complex
  independent component analysis,'' \emph{IEEE Trans. Signal Process.},
  vol.~61, no.~2, pp. 365--379, Jan. 2013.

\bibitem{Afsari_JointDiag_2008}
B.~Afsari, ``Sensitivity analysis for the problem of matrix joint
  diagonalization,'' \emph{SIAM J. Matrix Anal. Appl.}, vol.~30, no.~3, pp.
  1148--1171, Sep. 2008.

\bibitem{Comon_ICAConcept_1994}
P.~Comon, ``Independent component analysis, a new concept?'' \emph{Signal
  Process.}, vol.~36, no.~3, pp. 287--314, 1994.

\bibitem{Lavergne_CauchySchwarzIneq_2008}
P.~Lavergne, ``A {C}auchy-{S}chwarz inequality for expectation of matrices,''
  Department of Economics, Simon Fraser University, Discussion Papers, 2008.

\bibitem{Kim_ThesisIVA_2006}
T.~Kim, ``Independent vector analysis,'' Ph.D. dissertation, Department of
  BioSystems Korea Advanced Institute of Science and Technology, 2006.

\bibitem{Anderson_Kotz_2013}
M.~Anderson, G.-S. Fu, R.~Phlypo, and T.~Adal{\i}, ``Independent vector
  analysis, the {K}otz distribution, and performance bounds,'' in \emph{Proc.
  IEEE Int. Conf. Acoust., Speech Signal Process. (ICASSP)}, 2013, accepted.

\bibitem{Davies_AudioSourceSep_2002}
M.~Davies, ``Audio source separation,'' \emph{Math. Signal Process. V}, pp.
  57--68, 2002.

\bibitem{Petersen_MatrixCookboox_2008.pdf}
\BIBentryALTinterwordspacing
K.~B. Petersen and M.~S. Pedersen, ``The matrix cookbook,'' Nov. 2008.
  [Online]. Available: \url{http://matrixcookbook.com/}
\BIBentrySTDinterwordspacing

\bibitem{Muirhead_MultivariateStatisticalTheory_2005}
R.~J. Muirhead, \emph{Aspects of Multivariate Statistical Theory}.\hskip 1em
  plus 0.5em minus 0.4em\relax Wiley-In, 2005.

\bibitem{Aulogiaris_Kotz_2004}
G.~Aulogiaris and K.~Zografos, ``A maximum entropy characterization of
  symmetric {K}otz type and {B}urr multivariate distributions,'' \emph{TEST},
  vol.~13, pp. 65--83, 2004, 10.1007/BF02603001.

\end{thebibliography}
\end{document}